\pdfoutput=1
\documentclass[11pt]{article}

\usepackage[final]{acl}

\usepackage{times}
\usepackage{latexsym}

\usepackage[T1]{fontenc}
\usepackage[utf8]{inputenc}

\usepackage{microtype}

\usepackage{inconsolata}

\usepackage{graphicx}

\usepackage{booktabs}       % professional-quality tables
\usepackage{amsfonts}       % blackboard math symbols
\usepackage{nicefrac}       % compact symbols for 1/2, etc.
\usepackage{enumitem}

\usepackage[dvipsnames,table,xcdraw]{xcolor}

\usepackage{hyperref}

\usepackage{multirow}
\usepackage{wrapfig}

\usepackage{amsmath}
\usepackage{amssymb}
\usepackage{mathtools}
\usepackage{amsthm}

\newcommand{\draftonly}[1]{#1}

\newcommand{\draftcomment}[1]{\draftonly{#1}}
\newcommand{\todo}[1]{\draftcomment{\textcolor{red}{\small [TODO: #1]}}}

\newcommand{\ns}[1]{\draftcomment{\textcolor{blue}{\small [NPS: #1]}}}

\theoremstyle{plain}

\theoremstyle{definition}

\theoremstyle{remark}

\title{Sometimes I am a Tree:\\ Data Drives Unstable Hierarchical Generalization in LMs}
\author{
  Tian Qin \\
  Harvard University \\
   \texttt{tqin@g.harvard.edu} \\
   \And
  Naomi Saphra \\
  Harvard University \\
\texttt{nsaphra@fas.harvard.edu} \\
  \And
  David Alvarez-Melis \\
  Harvard University \& MSR \\
   \texttt{dam@seas.harvard.edu}  \\
}
\begin{document}

\maketitle




% keywords can be removed
% \keywords{First keyword \and Second keyword \and More}
% \input{sections/future_work}

\begin{abstract}
Early in training, LMs can behave like n-gram models, but eventually, they often learn tree-based syntactic rules and generalize hierarchically out of distribution (OOD).
We study this shift using controlled grammar-learning tasks: question formation and tense inflection. We find a model learns to generalize hierarchically if its training data is \textit{complex}---in particular, if it includes center-embedded clauses, a special syntactic structure. Under this definition, complex data drives hierarchical rules, while less complex data encourages shortcut learning in the form of n-gram-like linear rules. Furthermore, we find that a model uses rules to generalize, whether hierarchical or linear, if its training data is \textit{diverse}---in particular, if it includes many distinct syntax trees in the training set. Under this definition, diverse data promotes stable rule learning, whereas less diverse data promotes memorization of individual syntactic sequences. Finally, intermediate diversity and intermediate complexity form an \textit{unstable regime}, which is characterized by oscillatory learning dynamics and inconsistent behaviors across random seeds.
These results highlight the central role of training data in shaping generalization and explain why competing strategies can lead to unstable outcomes.
\end{abstract}

\section{Introduction}
 Early in training, LMs can behave like n-gram models, relying on local heuristics without capturing the deeper structure of language  \citep{Choshen2022-qj, Geirhos2020-ex, Saphra2018-xx}. However, they also exhibit breakthroughs in generalization by suddenly shifting into more sophisticated behaviors \citep{Choshen2022-qj, Chen2023-fi, McCoy2020-pj}. 
While previous works attribute these advanced capabilities to architecture and training objectives \citep{Ahuja2024-ul, McCoy2020-pj}, we investigate how two key \textit{data} characteristics---complexity and diversity---influence training outcomes.\footnote{Code can be found at \url{https://github.com/sunnytqin/hier_gen.git}.}

To study when models favor latent structures over surface heuristics, we focus on two controlled grammar-learning tasks: question formation and tense inflection \citep{McCoyUnknown-uy}. Each task can be solved either by a \textbf{linear rule}, which corresponds to an n-gram–like heuristic that operates on the most recent noun or auxiliary, or by the \textbf{hierarchical rule}, which reflects the correct syntactic structure used to generate the data. In the training and in-distribution (ID) data, sentences are constructed to be \textbf{ambiguous}, so both rules give the correct answer. In the out-of-distribution (OOD) data, this \textit{ambiguity is removed}, and only the hierarchical rule succeeds.

\begin{figure*}
    \centering
    \includegraphics[width=\textwidth]{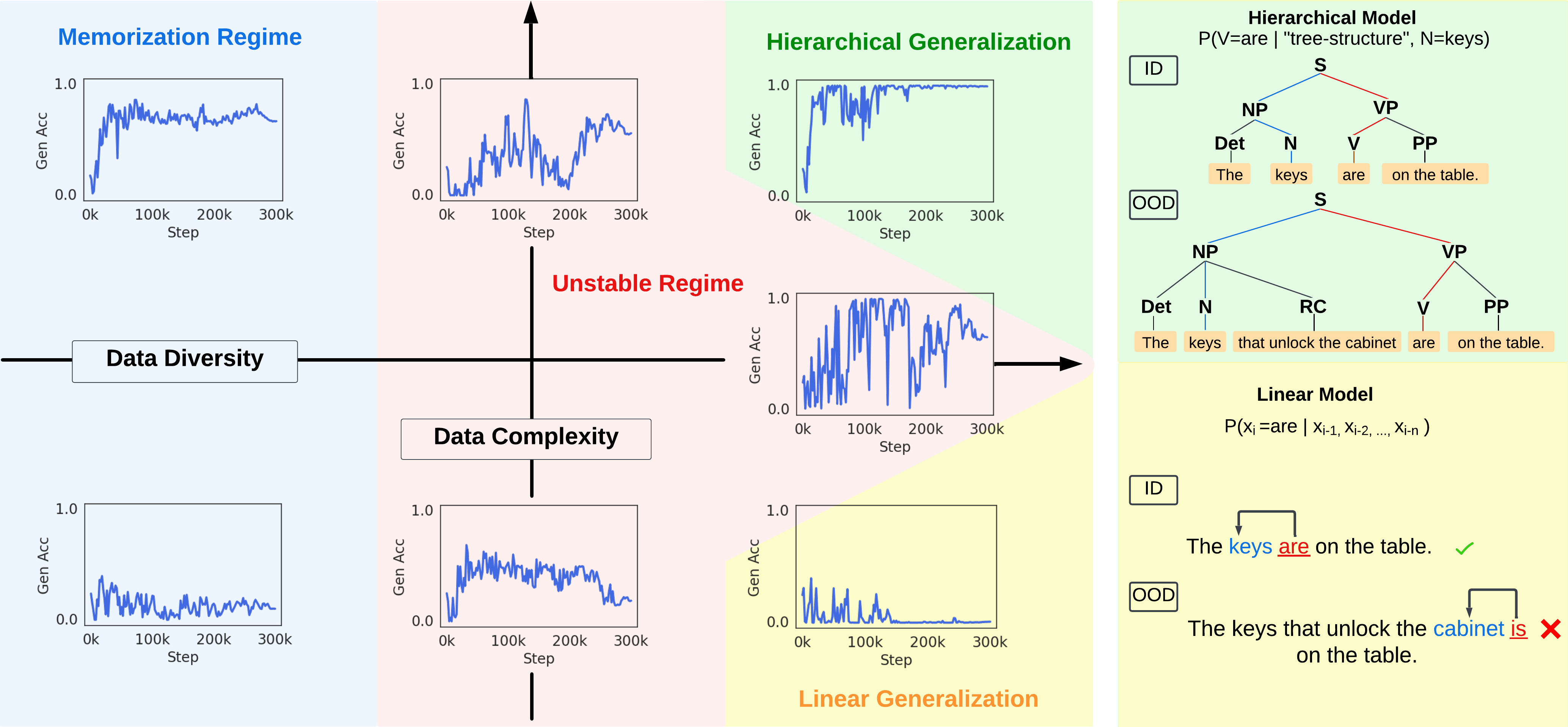}
    \caption{\textbf{Data plays a critical role in generalization behaviors and training stability.} 
    \textit{Left:} 
    Along the data diversity x-axis, low data diversity (as measured by variation in syntactic structure) leads the model to memorize unreliable sample-specific patterns, whereas high data diversity promotes commitment to a general rule. 
    Along the data complexity y-axis, high data complexity (as measured by the proportion of center-embedded sentences) induces the hierarchical rule, while simpler data (right-branching sentences) induces the surface-level linear rule. Mixing these data types results in unstable OOD training behaviors. 
    \textit{Upper Right:} A model that captures hierarchical structure of syntax can generalize grammatical rules OOD by correctly identifying the subject as the noun closest to the root on the syntax tree graph. 
    \textit{Lower Right:} A model that uses the linear rule will treat the most recent noun as the target verb's subject and thereby fail to generalize to unseen sentence compositions.
    } 
    \label{fig:stability_demo}
\end{figure*}

We illustrate the differences between rules in Fig.~\ref{fig:stability_demo}, where the model inflects the main verb of a sentence to agree with a subject's noun. In the n-gram-like linear rule (\textit{bottom right}), the model uses the nearest noun as the subject and fails when a distractor noun appears in a prepositional phrase. In the hierarchical case \textit{(top right}), the model uses a latent tree structure to correctly identify the subject and apply, enabling generalization to any grammatical sentence. Prior work shows that transformer-based LMs can shift from the linear heuristic to the hierarchical rule when trained long enough, a transition called \textbf{structural grokking} \citep{Murty2023-xp}, in parallel to the well-known shift from memorization to generalization in classic grokking \citep{Power2022-hz}.

Building on this line of work \citep{McCoy2018-uv, McCoy2020-pj, Ahuja2024-ul, Murty2023-xp}, we study how training \textit{data} properties shape whether models adopt the hierarchical rule or default to the linear rule. We define \textbf{data complexity} as the presence of center-embedded clauses, a syntactic structure where the subject is interrupted by a relative clause. Our experiments show that such complex structures are the key driver of hierarchical OOD generalization; models trained without them consistently fall back on shortcuts. This finding echoes a classic claim in linguistics \citep{wexler1980formal} that center embeddings play a central role in human syntax acquisition.

After identifying center embeddings as the driver of hierarchical generalization, we next ask how data composition influences rule competition during training. \citet{Ahuja2024-ul} showed that linear and hierarchical rules can coexist and compete during learning. We show that the outcome of this competition depends on the training data. While in-distribution accuracy is always stable across time and consistent across seeds, OOD accuracy is often \textit{unstable} during training and \textit{inconsistent} across seeds. The two rules compete, and only runs that fully commit to one rule---linear or hierarchical---exhibit stable OOD performance. These commitments are determined by two data  properties: \textbf{complexity} (presence of center embeddings) and \textbf{diversity} (number of distinct syntax trees). We find that commitment consistently occurs only under both high complexity and high diversity, leading to stable generalization. At intermediate levels of either  property, models fail to commit, resulting in unstable training and inconsistent outcomes across seeds. As summarized in Fig.~\ref{fig:stability_demo}, diversity promotes general rules over memorization, while complexity determines which rule is learned.

Our controlled synthetic settings allows us to precisely manipulate data properties and isolate their effects on rule learning. The resulting insights extend beyond grammar learning: they highlight how data complexity and diversity govern whether models rely on memorization, heuristics, or systematic generalization. These dynamics connect to broader themes in the study of grokking and training instability \citep{Power2022-hz, Ahuja2024-ul}, random variation between training runs \citep{Juneja2022-hj, Hu2023-vh}, and phase transitions in neural networks \citep{Schaeffer2023-od, Theunissen2020-pv}. The mechanisms we identify in a simplified domain may inform our understanding of generalization in larger-scale models. Our contribution can be summarized as follows:
\begin{itemize}[itemsep=2pt,labelindent=2pt,topsep=0pt,parsep=0pt,partopsep=1pt, align=left, leftmargin=*]
    \item We show that data complexity---that is, center embedding frequency---drives models to adopt hierarchical syntax rather than n-gram heuristics (Section~\ref{sec:data_complexity}).
    \item We find that models only achieve stable OOD performance when they commit to a general rule, whether linear or hierarchical. Competing rules lead to unstable training and inconsistent outcomes across seeds (Section~\ref{sec:stability}).
    \item We show that data diversity separates memorization, instability, and generalization regimes (Section~\ref{sec:data_diversity}).

\end{itemize}

\begin{table*}[t]
    \centering
    {\renewcommand{\arraystretch}{1.1}
    \small
      \caption{\textbf{Examples from two grammar case studies.} \textit{Top}: In the question formation task, the model moves the main auxiliary verb to the front to form a question.
      \textit{Bottom}:  In the tense inflection task, the model inflects the main verb from past to present tense, while respecting subject-verb agreement. }  
    \label{tab:task_examples}
    \resizebox{\textwidth}{!}{
    \begin{tabular}{p{2.5cm} | p{2.3cm} p{9cm}}
    \hline
    \textbf{Dataset} & \textbf{Rule Type} & \hspace{3.5cm} \textbf{Examples} \\ 
    \hline
    \multirow{2}{*}{Question Formation} & Ambiguous  & \textbf{Input:} My unicorn \textcolor{ForestGreen}{does} move the dogs that \textcolor{red}{do} wait.  \\ 
                             &  &\textbf{Output:} \textcolor{ForestGreen}{Does} my unicorn move the dogs that \textcolor{red}{do} wait?     \\
                             \cline{2-3} 
                             & \multirow{2}{*}{Unambiguous }   &\textbf{Input:} My unicorn who \textcolor{red}{doesn't} sing \textcolor{ForestGreen}{does} move.
 \\ 
                             & & \textbf{Linear Output:} \textcolor{red}{Doesn't} my unicorn who sing \textcolor{ForestGreen}{does} move?
 \\
                             & & \textbf{Hierarchical Output:} \textcolor{ForestGreen}{Does} my unicorn who \textcolor{red}{doesn't} sing move? \\
                              
                             % & \multirow{2}{*}{Decl}   & \textbf{Input:} My unicorn does move the dogs that do wait. \\ 
                             % & & \textbf{Output:} My unicorn does move the dogs that do wait.    \\
    \hline
    \multirow{5}{*}{Tense Inflection} & Ambiguous & \textbf{Input:} My zebra behind the peacock smiled. \\ 
                        &  & \textbf{Output:} My \textcolor{cyan}{zebra} behind the \textcolor{cyan}{peacock}  \textcolor{ForestGreen}{smiles}.    \\
                        \cline{2-3} 
                        & \multirow{1.5}{*}{Unambiguous}   & \textbf{Input:} My zebra behind the peacocks smiled. \\ 
                        & & \textbf{Linear output:} My zebra behind the \textcolor{cyan}{peacocks} \textcolor{red}{smile}.     \\
                        &   & \textbf{Hierarchical output:} My \textcolor{red}{zebra} behind the peacocks \textcolor{ForestGreen}{smiles}.     \\
                        % & \multirow{2}{*}{Decl}   & Input: My unicorn does move the dogs that do wait. \\ 
                        % & & Output: My unicorn does move the dogs that do wait.    \\
    \hline
    \end{tabular}
    \vspace{-5px}
    }}
\end{table*}

\section{Experimental Setup}
\label{sec:experiments}
The question formation and tense inflection tasks were first proposed by \citet{Frank2007-pn} and \citet{Linzen2016-vx} as canonical tests of language modeling ability. We use existing synthetic datasets for question formation from \citet{McCoy2018-uv} and tense inflection from \citet{McCoy2020-pj}.
\subsection{Question Formation Task}
\label{sec:qf_task}

In the \textbf{question formation (QF)} task, the model transforms a declarative sentence into a question by moving the main auxiliary verb (such as \textit{does} in \textit{does move}) to the front. Our training data (based on \citet{McCoy2018-uv}) permits two strategies for choosing which verb to move: (1) a linear rule that moves the first auxiliary verb (Fig.~\ref{fig:stability_demo} \textit{top right}), or (2) a hierarchical rule---the correct rule in English grammar---based on the sentence's syntax tree (Fig.~\ref{fig:stability_demo} \textit{bottom right}), which places the main auxiliary verb above other verbs.

Examples of each rule are provided in Table~\ref{tab:task_examples}. The first example is considered \textit{ambiguous} because the hierarchical and linear rules produce the same correct outcome. In contrast, the second example is \textit{unambiguous} because only the hierarchical rule produces the correct outcome. The training and in-distribution test data contain only ambiguous samples, while the OOD generalization set includes only unambiguous samples. Therefore, if a model uses the hierarchical rule, it will achieve $100\%$ accuracy on both the in-distribution (ambiguous questions) and OOD (unambiguous questions) sets. Conversely, if a model uses the linear rule, it will still score 100\% accuracy on the in-distribution set, but $0\%$ on the OOD set. 
We therefore use the model's accuracy on the OOD set to measure hierarchical generalization.

\subsection{Tense Inflection Task}
\label{sec:ti_task}

In the \textbf{tense inflection (TI)} task, the model transforms a past-tense sentence into the present tense by changing the inflection of its main verb. 
In English, past-tense verbs (\textit{walked}) are not inflected by plurality, so the model must identify the subject and use its plurality to inflect the present-tense verb (\textit{walk}/\textit{walks}). The TI training data again follows either a hierarchical or linear rule for subject-verb agreement (based on \citet{McCoy2020-pj}). The linear rule inflects the verb based on the most recent noun, while the hierarchical rule correctly inflects according to the subject. As in the QF task, the training and ID validation sets contain ambiguous examples, whereas the OOD set contains unambiguous examples. In the ambiguous example from Table~\ref{tab:task_examples}, the subject noun \textit{zebra} and the most recent noun \textit{peacock} share the same plurality, so either rule produces the correct answer. In the OOD unambiguous example, the subject and the most recent noun differ in plurality and therefore only the hierarchical rule produces the correct answer.
Similar to the QF task, we use the model's main-verb prediction accuracy on the OOD set as a metric for hierarchical generalization.

\begin{figure*}[t!]
    \centering
    \includegraphics[width=1.0\textwidth]{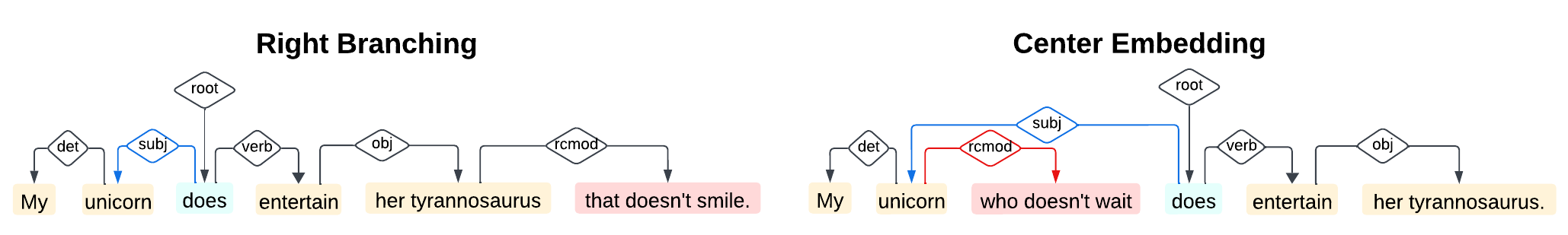}
    \caption{\textbf{Sentence Examples.}   \textit{Left:} Right-branching sentence example. The linear progression of the main constituent is not interrupted by the relative clause. 
    \textit{Right:} Center-embedded sentence example. When the relative clause modifies the subject, it interrupts the linear progression of the main constituent. 
    }
    \label{fig:sentence_demo}
\end{figure*}
\subsection{Models, Data and Training}
\label{sec:model_and_training}
We run all experiments on the same 50 random seeds using hyperparameter settings from the existing literature \citep{Ahuja2024-ul, Murty2023-xp}. We use a decoder-only Transformer architecture where each layer has 8 heads with a 512-dimensional embedding. QF models have 6 layers and TI models have 4 layers. All models are trained from scratch on a causal language modeling objective for 300K steps. We use the Adam optimizer \citep{Kingma2014-he}, a learning rate of 1e-4, and a linear decay schedule. We use a word-level tokenizer with a vocabulary of size 72. 

We use the original training, validation and OOD test data proposed by \citet{McCoy2018-uv} and \citet{McCoy2020-pj}. Where we expand the training data for our data composition experiments, we mimic the data generation process used for the original QF and TI task. Specifically, the original TI and QF data are generated with Context-Free Grammars (CFGs) using a simplified set of grammatical rules; we reuse the same CFG rules to create variations of the training data.

% \section{Center Embedding Leads to The Hierarchical Rule}
\section{Data Complexity Determines the Rule}
\label{sec:data_complexity}

We begin by examining how the complexity of training data influences the rules that models learn. Our focus is on center-embedded sentences, a syntactic structure that has long been used in linguistics to characterize complexity. By ablating center embeddings from training data, we test how data complexity shapes which rule a model learns.

\subsection{Center Embedding}
\label{sec:center_embed}
Center embedding occurs when a clause is placed recursively within another clause of the same type. Fig.~\ref{fig:sentence_demo} illustrates two examples of center-embedded sentences, where the embedded clause complicates syntactic parsing by placing an additional subject noun in between a verb and its own subject. 

Without center embeddings, English syntax only branches recursively to the right. In exclusively right-branching structures, modifying clauses are always appended at the end of the main clause, maintaining linear flow. Therefore, linguists have long argued that center embeddings play a crucial role in grammar acquisition \citep{wexler1980formal} and lead to tree-like syntactic structures \citep{Chomsky2015-bg}. 
We find center embeddings, which are crucial for human language acquisition, also lead an LM to prefer hierarchical grammar rules. 

Linguistic theory explains that center embeddings lead to hierarchical rule learning because of their greater computational requirements. To predict the next token, LMs must track syntactic connections between words in the context. In right-branching sentences, LMs can rely on linear proximity to identify these connections; as shown in Fig.~\ref{fig:sentence_demo}, a simple bigram model suffices to capture the subject-verb relationship for such sentences. In contrast, center embeddings introduce relative clauses of various lengths, making linear n-gram models inefficient for capturing subject-verb relationships. Because center embeddings are recursive, they require the model to track multiple subject-verb relationships: one for the main clause and a separate one for the embedded relative clause. In these cases, a tree structure is more efficient than a linear one to model subject-verb relationships.

\begin{figure*}[t!]
    \centering
    % \includegraphics[width=0.38\linewidth]{figures/no_curriculum_main.pdf}
    % \hfill
    % \includegraphics[width=0.5\linewidth]{figures/ti_simplicity_contamination.pdf}
    \includegraphics[height=0.3\linewidth]{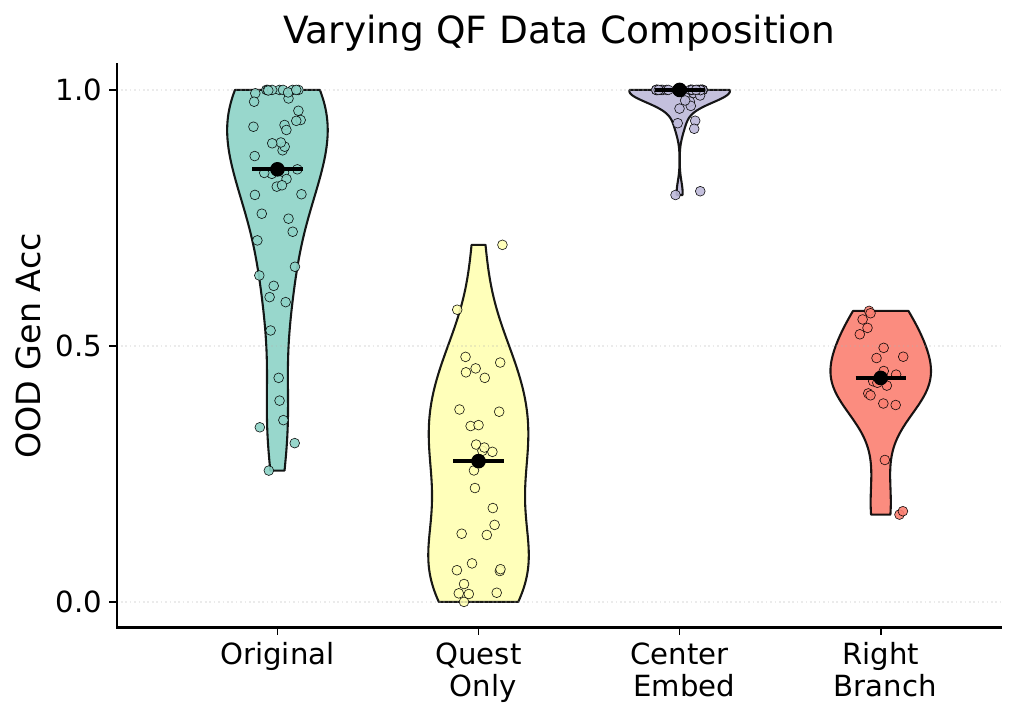}
    \includegraphics[height=0.3\linewidth]{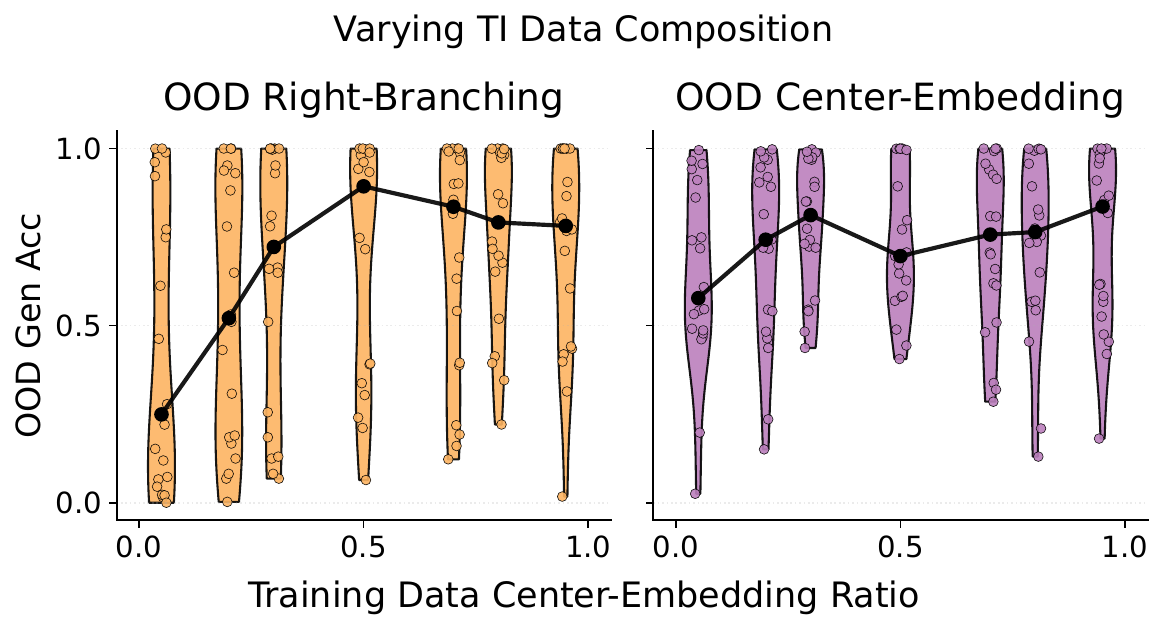}
    \caption{
    \textbf{Components of training data drive different generalization behaviors.} Each dot represents one model trained from a different random seed.
    \textit{Left:} In QF, center-embedded sentences appear only in declaration-copying data; exposure to these examples induces hierarchical generalization.
    \textit{Right:} In TI, we vary the mix of right-branching and center-embedded sentences and evaluate OOD performance on right-branching (\textit{orange}) vs. center-embedded (\textit{purple}).}
    \label{fig:grokking_selection}
\end{figure*}
\subsection{Question Formation Results}
\label{sec:qf_result}
In the QF task  (Section~\ref{sec:qf_task}), the training data must remain ambiguous between the linear rule (moving the first auxiliary) and the hierarchical rule (moving the main auxiliary). Because center-embedded sentences are not ambiguous, they cannot appear in QF training examples. To expose the model to such structures, \citet{McCoy2018-uv} introduced a secondary task: declaration copying. Like QF, it begins with a declarative sentence, but instead of transforming it, the model simply repeats it. Only the primary QF task enforces ambiguity, so declaration-copying examples may include center embeddings. In our first set of experiments, we modify the composition of the declaration-copying subset to greate three new training sets: \textit{Quest Only} (no declaration-copying examples), \textit{Center Embed} (only center-embedded declarations), and \textit{Right Branch} (only right-branching declarations). In all cases, the full set of QF training examples is retained. We train models using 50 random seeds.

Regardless of training set composition, all models achieve 100\% accuracy on the in-distribution test data. However, their OOD performance differs sharply (Fig.~\ref{fig:grokking_selection}, left). Models trained without any center-embedded declaration-copying examples universally (across all 50 seeds) fail to learn the hierarchical rule, even when right-branching declarations are included. By contrast, models trained with only center-embedded declaration-copying examples strongly favor the hierarchical rule. These results demonstrate that exposure to center-embedded sentences is essential for inducing hierarchical generalization.

\subsection{Tense Inflection Results}
\label{sec:ti_result}

In the TI task (Section \ref{sec:ti_task}), both right-branching and center-embedded sentences are ambiguous as long as the verb's subject shares the same plurality as the distractor noun between the subject and verb. For right-branching sentences, the distractor occurs in a prepositional phrase; for center-embedded sentences, it occurs in a relative clause. We show two concrete examples below.

\begin{enumerate}[itemsep=3pt,labelindent=0pt,topsep=0pt,parsep=0pt,partopsep=1pt, align=left, leftmargin=*]
    \item \textbf{Right Branching}: The noun (\textit{cabinet}) in the prepositional phrase (\textit{to the cabinet}) acts as the distractor for the TI task.
    
    ID: \textit{The \underline{keys} to the \textbf{cabinets} \underline{are} here.}

    OOD: \textit{The \underline{keys} to the \textbf{cabinet} \underline{are} here.}
    
    \item \textbf{Center Embedding}: Either the subject or the object (\textit{cabinet}) in the relative clause (\textit{that unlock the cabinet}) acts as the distractor.

    ID: \textit{The \underline{keys} that unlock the \textbf{cabinets} \underline{are} here.}

    OOD: \textit{The \underline{keys} that unlock the \textbf{cabinet} \underline{are} here.}

\end{enumerate}

To test the effect of center embeddings on rule learning, we create training sets that vary the ratio of right-branching to center-embedded sentences from 5\% to 95\%, while keeping the total number of samples fixed.\footnote{The original dataset also included a secondary past-tense copying task, parallel to declaration copying in QF. We show in App.~\ref{appdx:ti_secondary} that this task is not necessary and omit it here.} For evaluation, we split the original OOD test set \citep{McCoy2020-pj} into right-branching and center-embedded subsets. This split ensures we can separately measure how well models generalize on each kind of tree.

All seeds achieve perfect accuracy on the in-distribution test set, but OOD performance depends strongly on the proportion of center-embedded sentences (Fig. \ref{fig:grokking_selection}, right). On the least complex training dataset, where only 5\% of sentences have center embeddings, most seeds achieve below-random performance on the right-branching OOD subset and around random on the center-embedded subset. As the proportion of center embeddings increases (up to 50\%), the majority of seeds succeed on both OOD subsets, indicating that models have learned the hierarchical rule. These findings confirm that increasing data complexity through center embeddings promotes hierarchical generalization. In Appendix \ref{appdx:obj_sbj_ctr_breakdown}, we further show that different subtypes of center embedding vary in their effect on TI rules.

\begin{figure*}[t]
    \centering
    \includegraphics[width=0.7\linewidth]{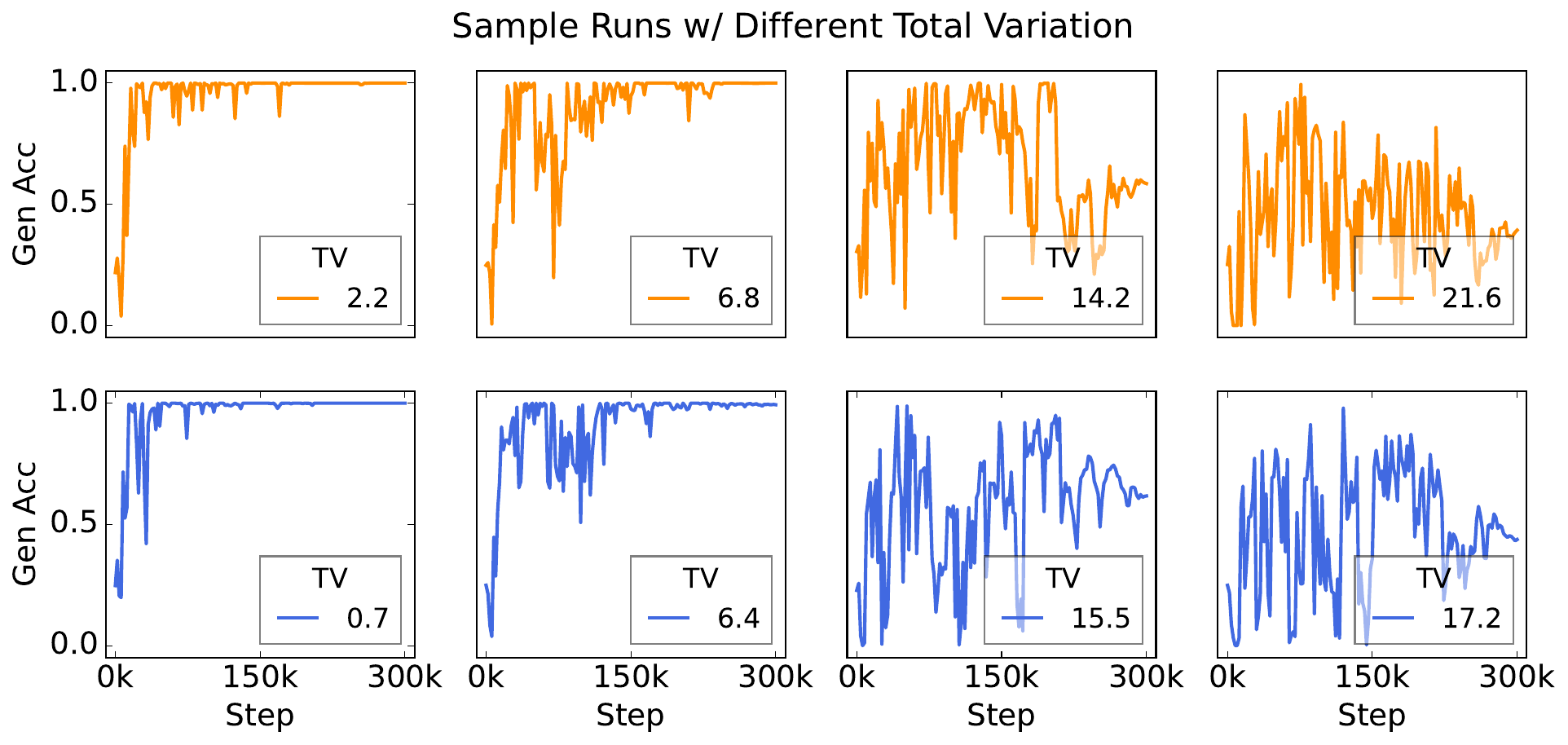}
    % \caption{Sample Runs with different total variation}
    \caption{\textbf{Each training run either stabilizes in a simple OOD generalization rule or oscillates in its OOD accuracy.} The OOD generalization behaviors can be either stable or unstable  when trained on different seeds. We use total variation to quantify the instability within one training run.}
    \label{fig:sample_runs}
\end{figure*}
\section{Rules stabilize training}
\label{sec:stability}
Why do some runs fail to reliably generalize hierarchically even when trained on hierarchy-inducing data? This section will show that these poor runs are oscillating between competing rules; models only stabilize OOD if they commit to a general rule, whether hierarchical or linear. Under our training conditions, some random seeds fail to stabilize regardless of training set composition.

\subsection{Instability During Training} 
\label{sec:tv_def}
Across all our model runs, training loss and ID performance are both consistent across random seeds and stable during training (See ~\ref{appdx:training_instabilty} for example training curves). Despite stable ID accuracy during training, some random seeds lead to highly unstable OOD accuracy accuracy oscillating during training. We measure OOD instability across training time using a standard metric from signal processing: \textbf{total variation} (TV). Specifically, we checkpoint the model every 2K steps and measure the generalization accuracy $\mathrm{Acc}_t$ at each checkpoint timestep $t \in T$. TV is defined as: 
% \begin{wrapfigure}[]{r}{0.4\textwidth}
\begin{align}
    \text{TV} = 
    \frac{1}{|T|} \sum_{t \in T} 
    \left| \mathrm{Acc}_t - \mathrm{Acc}_{t-1} \right|
\end{align}
In Figure \ref{fig:sample_runs}, we show the OOD performance of four runs trained on different seeds - ranging from highly stable OOD performance to high unstable ones. We also report corresponding TV to demonstrate that TV quantifies training instability.

\subsection{Stability Ties to Rule Commitment}
\label{sec:intra_inter}
Why do some runs exhibit stable OOD performances while others oscillate wildly? 
We now demonstrate that regardless of training data mix, training runs with stable OOD behavior  always commit to a universal systematic rule. 

\begin{figure*}[t!]
    \centering
    \includegraphics[width=0.9\linewidth]{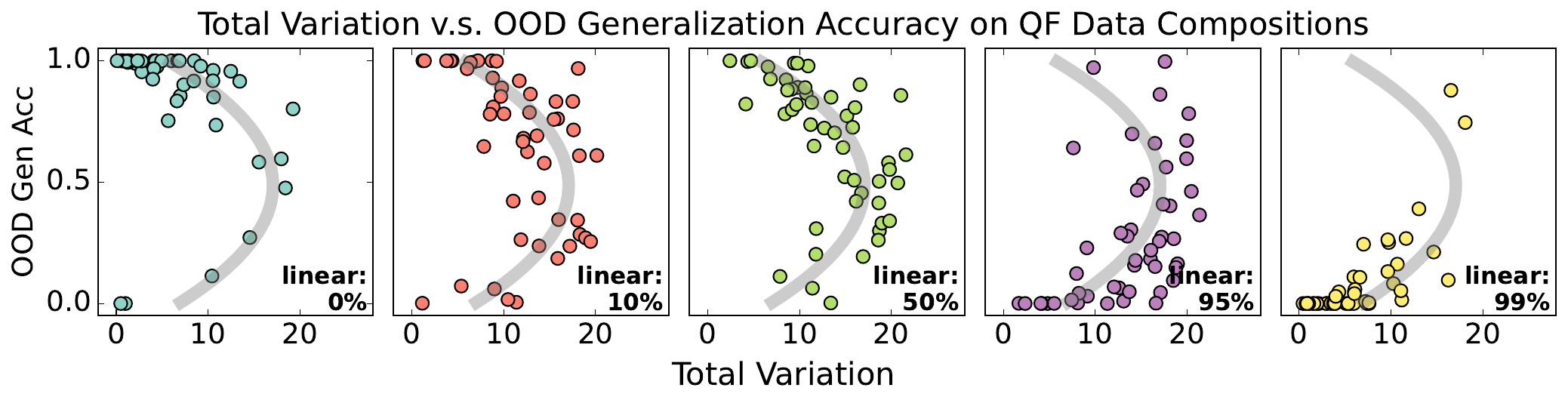}
    \caption{\textbf{Training stability vs. final generalization accuracy for QF task.} We create heterogeneous training data by mixing different proportions of linear-inducing data and hierarchical-inducing data. For each data mix, we train models on 50 random seeds and for each seed, we examine training instability ($x$-axis) and final OOD generalization performance ($y$-axis). Grey line indicates the smoothed average curve across all five datasets.
    }
    \label{fig:intra_inter_variance}
    % \vspace{-5px}
\end{figure*}

\paragraph{Setup: }
We construct five variations of the training data using the following procedure. Each new dataset contains 50K questions from the original data and 50K declarations with a controllable ratio between center-embedded (hierarchy-inducing) and right-branching (linearity-inducing) sentences. Our newly generated declarations maintain the original dataset's distribution of unique syntax trees. Specifically, for each sentence in the original data, we keep its CFG-based syntax tree but resample words from the CFG's vocabulary. We then train models from 50 random seeds using the hyperparameters from Section \ref{sec:model_and_training}. For each seed, we report model OOD performance and TV in Figure~\ref{fig:intra_inter_variance}. 

% By controlling for training instability, we reveal that generalization behavior is clustered and highly bimodal across random seeds. 
\paragraph{Results: } As shown in Figure~\ref{fig:intra_inter_variance}, regardless of the data mix, the final OOD generalization accuracy for all stable models is either $100\%$ or $0\%$---that is, they all commit to a universal systematic rule. 
While either rule can be implemented by a stable model, training data composition determines the likelihood of training outcomes---whether a run  stabilizes and whether its systematic rule is  hierarchical or linear. As seen previously, if the training data is dominated by either center-embedded or right-branching sentences, the resulting models usually commit to the hierarchical or linear rule, respectively. 
However, when the data is heterogeneous (e.g., 10\% of examples are linearity-inducing right-branching sequences), the final generalization accuracy of stable runs is bimodally distributed across random seeds, clustering around $100\%$ or $0\%$. 
 In fact, the horseshoe-shaped curves in Figure~\ref{fig:intra_inter_variance} illustrate that the less stable a QF training run is, the less universally the model applies its OOD rules. (See Appendix~\ref{appdx:tense_tv} for matching results on the TI task.) 

In summary, by mixing linear-inducing and hierarchical-inducing data, we create heterogeneous training data mixture. When training on heterogeneous data mixtures, more seeds lead to unstable runs. However, even with heterogeneous mixes, some runs can still stabilize by committing to one of the competing rules.

\section{Data Diversity Leads to General Rules}
\label{sec:data_diversity}
In the previous section, we showed that training runs stabilize if the model commits to a systematic rule. We now extend this finding by showing that data diversity promotes commitment to these systematic universal rules. We find that training can stabilize in two ways: with \textit{less} diverse data, models stabilize by memorizing training examples, while with \textit{more} diverse data, they stabilize by committing to systematic rules (studied in the previous section). At intermediate diversity levels---as at intermediate complexity levels (Section \ref{sec:intra_inter})---training becomes unstable. 

\subsection{Measuring Data Diversity}
\label{sec:data_diveristy}

We define a dataset's diversity according to the syntactic similarity between its sentences. 
We measure a sentence pair's similarity by the tree-edit distance (TED) between their respective latent syntax trees \citep{Chomsky2015-bg}. When two sentences share the same syntax tree, we can transform one into the other while only changing their leaf-node vocabulary. For example, \textit{My unicorn loves her rabbit} and \textit{Your zebra eats some apples} have different vocabulary but identical syntax trees. We define a dataset's \textbf{diversity} as the number of unique syntactic trees it contains, similar to diversity metrics previously used in both natural language \citep{Huang2023-ab, Gao2024-fi, Ramirez2022-mx} and code \citep{Song2024-cg}. 

\subsection{Diversity and stability}
\label{sec:inverse}

In Section \ref{sec:stability}, we showed that training stabilizes if models commit to a systematic rule. Here, we show that data diversity shapes training instability and rule commitment. Specifically, we find that models trained on less diverse data can stabilize through memorization without committing to any systematic rules, whereas those trained on more diverse data stabilize through systematic rules. Previously, we found that \textit{which} rule a model commits to depends on whether the training set is dominated by hierarchy-inducing data or linearity-inducing data, so we will analyze these two settings separately.

\paragraph{Hierarchy-inducing data}  
To study the effect of diversity in hierarchy-inducing settings, we construct multiple training sets with varying levels of syntactic diversity. Each set contains 50K question samples and 50K center-embedded declarations. We control diversity by changing the number of unique syntactic trees in the declarations. For each dataset, we train 50 random seeds and measure instability using total variation (Section~\ref{sec:tv_def}). We also report a \textbf{rule commitment ratio}: the proportion of runs achieving OOD accuracy either above 95\% or below 5\%, which both indicate that the model has committed to a systematic rule.

\begin{figure*}[t]
    \centering
    \includegraphics[width=0.4\textwidth]{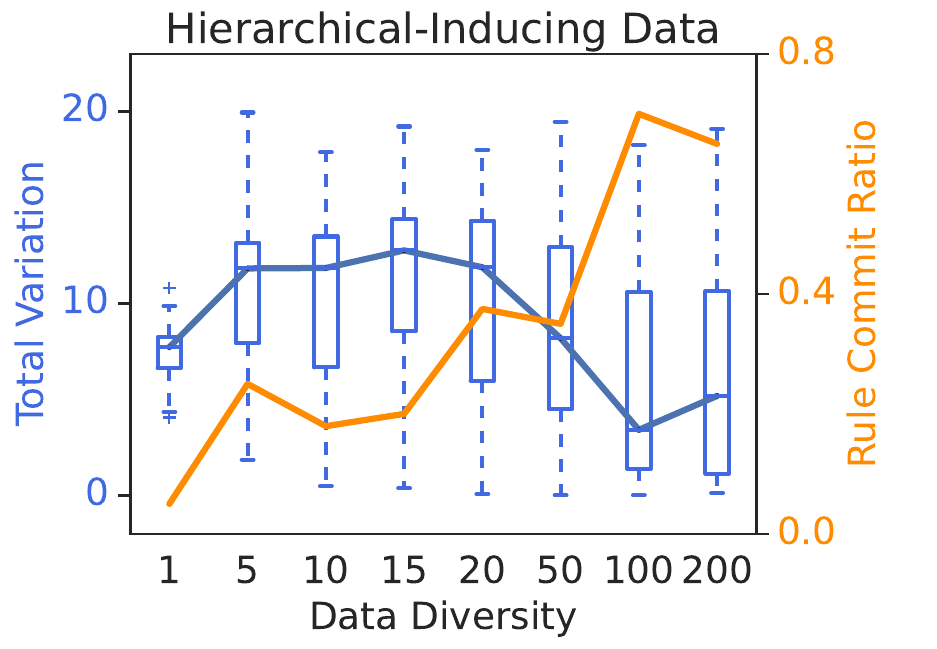}
    \includegraphics[width=0.4\textwidth]{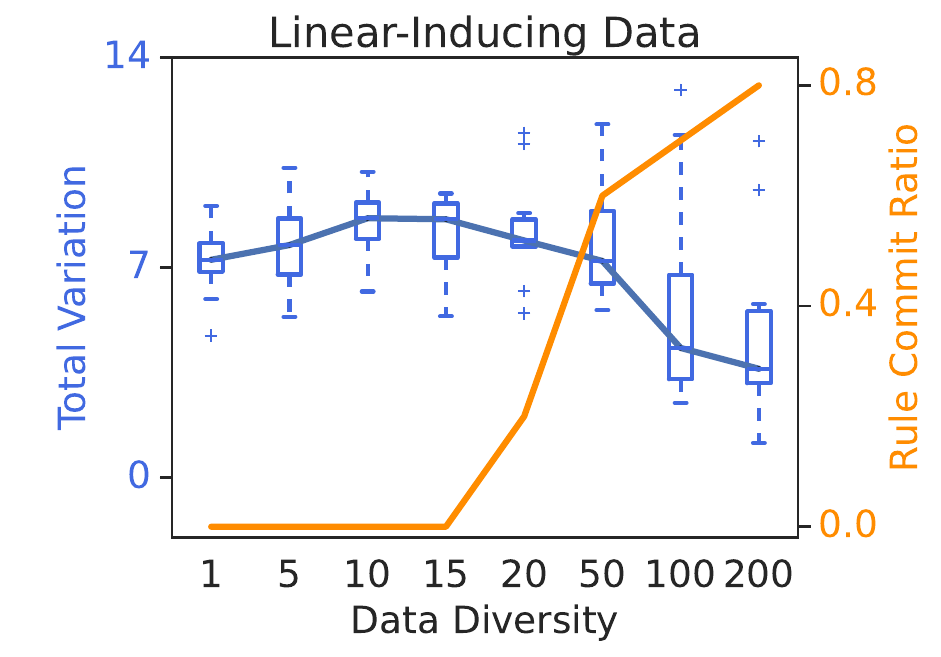}
    \caption{\textbf{Effect of data diversity on training stability and rule commitment.} The x-axis measures syntactic diversity as the number of unique syntax trees in the training set (Section XX). The left y-axis shows training instability, measured with total variation (Section~\ref{sec:tv_def}). The right y-axis shows the rule commitment ratio, defined as the proportion of 50 training seeds that achieve OOD accuracy either above 95\% or below 5\%, indicating commitment to a systematic rule. \textit{Left}: hierarchy-inducing data; \textit{Right}: linearity-inducing data, where diversity is varied through the small fraction of hierarchical examples. In both settings, low diversity leads to stable memorization, intermediate diversity to instability, and high diversity to stable rule commitment.} 
    \label{fig:data_diversity_uscale}
\end{figure*}

Figure~\ref{fig:data_diversity_uscale} (\textit{left}) shows three distinct regimes. At low diversity (diversity $<5$), models enter the \textbf{memorization regime}: training is stable, but they fail to commit to a rule. In Appendix~\ref{appdx:memorizaition}, we confirm that these memorizing models apply the hierarchical rule to syntactic structures seen during training, but cannot extrapolate to unseen structures. At high diversity (at least 50 unique trees), models enter the \textbf{hierarchical generalization regime}, where training reliably stabilizes by committing to the hierarchical rule. Between these extremes lies an \textbf{unstable regime} (5--20 unique trees), where the data is too diverse to memorize but not diverse enough to teach a universal rule, leading to oscillating OOD curves.

\paragraph{Linearity-inducing data} 
Unlike the hierarchy-inducing setting, linearity-inducing data (right-branching sentences) permits little syntactic variation: the auxiliary always follows the subject noun, leaving no natural way to increase diversity. Instead, we build on a mix of 99\% right-branching and 1\% center-embedded data, which we find reliably induces the linear rule. Keeping this mixture ratio, we control diversity by varying the syntactic structures of the center-embedded sentences. As in the hierarchy-inducing setting, we use sets of 50K questions and 50K declarations and measure training instability and rule commitment as before. 

Figure~\ref{fig:data_diversity_uscale} (\textit{right}) shows that the same three regimes appear in the linearity-inducing case. At low diversity, models memorize familiar syntactic structures without committing to a rule. At intermediate diversity, training becomes unstable; memorization is no longer possible but the signal is too weak for to induce a systematic generalization rule. At high diversity, models stabilize in the \textbf{linear generalization regime}, committing to the linear rule. This mirrors the U-shaped pattern observed with hierarchy-inducing data, confirming that syntactic diversity is essential for systematic generalization---regardless of which rule is preferred.

\section{Related Work}

See App.~\ref{appdx:related} for a more detailed literature review.

\paragraph{Syntax and Hierarchical Generalization}
\label{sec:syntax_related}
% \paragraph{Syntax and Hierarchical Generalization}
\citet{McCoy2018-uv} first used the question formation task to study hierarchical generalization in neural networks, showing that attention mechanisms improved generalization performance in recurrent neural networks (RNNs). Later, \citet{McCoy2020-pj} found that tree-structured architectures consistently induce hierarchical generalization. \citet{Petty2021-pe} and \citet{Mueller2022-rm} further concluded that  transformers tend to generalize linearly. This view was challenged by \citet{Murty2023-xp}, who attributed their results to insufficient training: decoder-only transformers can generalize hierarchically, but only after in-distribution performance has plateaued. They named this transition from surface-level heuristics to hierarchical generalization structural grokking. Expanding on their findings, \citet{Ahuja2024-ul} showed that models only generalize hierarchically when trained on a language modeling objective. All of this prior work attributed hierarchical inductive bias to model architecture or objective, whereas our study highlights the impact of data. 
While previous work observed some inconsistency across seeds \citep{McCoy2018-uv, McCoyUnknown-uy}, we characterize this random variation more specifically.

\paragraph{Training Dynamics and Grokking}
During \textit{grokking}, a neural network generalizes to a test set long after it has overfitted to its training data. \citet{Power2022-hz} first observed this phenomenon in simple arithmetic tasks. 
This \textit{classic} grokking is different from our main focus of \textit{structural} grokking \citep{Murty2023-xp}. In classic grokking, the model transitions from memorization to generalization, allowing it to achieve non-trivial performance on unseen data from the same distribution as the train set. In structural grokking, a model transitions from the simple linear rule to the hierarchical rule, allowing non-trivial performance on OOD data. However, our findings also relate to classic grokking by connecting data diversity to memorization. 

\citet{Zhu2024-nz} studied the role of data and found that grokking only occurs when training set is sufficiently large, and thus more diverse. \citet{Berlot-Attwell2023-qx} studied how data diversity leads to OOD compositional generalization in multimodal models and 
\citet{Lubana2024-ed} showed that diversity also induces compositional behaviors late in LM training.
\citet{Liu2022-mj} showed grokking can be induced by forcing a specific weight norm, a measurement of model---though not data---complexity. 
\citet{Huang2024-aw} and \citet{Varma2023-iq} have shown that different circuits compete during training and that different data and model sizes lead to different competition and training dynamics. Competition also shapes other phase transitions, such as transient in-context learning \citep{Park2024-ri}.
% While these works primarily study training dynamics, our findings highlight that this competition can also lead to inconsistent outcomes when training converges.
% We unify these threads in the existing grokking literature by characterizing the unstable regime in both data diversity and data complexity, connecting training stability with consistency under random seeds.

\paragraph{Random Variation} 
Although choices like hyperparameters, architecture, and optimizer all shape model outcomes, training remains inherently stochastic. Models are sensitive to random initialization and the order of training examples \cite{Dodge2020-pb,Zhou2020-xt, D-Amour2022-tl, Naik2018-og,Sellam2021-rz,Juneja2022-hj}. On text classification tasks, \citet{Zhou2020-xt} observed OOD variability  even late in training. We investigate the source of these training inconsistencies and link them more precisely to characteristics of the training data. 
% Our extended literature review in Appendix~\ref{appdx:related} expands on the following background overview. 

\section{Discussion and Conclusions}

By exploring the role of data in OOD generalization rules, we have also revealed which settings render outcomes unpredictable. Complex data induces more complex rules, but a mix of complex and simple examples leads to instability and inconsistency. Likewise,  higher data diversity favors universal generalization rules over example memorization, but intermediate diversity leads to instability and inconsistency. These findings have implications across machine learning and linguistics.

\paragraph{Inconsistent behavior across seeds.} While variation in model error is often treated as unimodal Gaussian noise in the theoretical literature \citep{Lakshminarayanan2016-fb}, our findings suggest that errors may only be distributed unimodally for a given compositional solution. Our work joins the growing literature that suggests random variation can create clusters of OOD behaviors. Previously, clusters were seen in text classification heuristics \citep{Juneja2022-hj} and training dynamics \citep{Hu2023-vh}. In our case, OOD accuracy is clearly bimodal only when we exclude unstable training runs. We suggest that research on variation in training consider training stability in the future, which may expose clustered solutions.

\paragraph{Implications for formal linguistics.}
% The learning theory field, although now seen primarily as an area of statistical machine learning, originated from formal linguistics. 
In debates about the poverty of the stimulus, linguists have extensively studied the question of what data is necessary and sufficient to learn grammatical rules \citep{McCoy2018-uv, Berwick2011-xb}. In particular, \citet{wexler1980formal} argued that all English syntactic rules are learnable given ``degree 2'' data: sentences with only one embedded clause nested within another clause. Our center embedding results confirm that without a stronger architectural inductive bias---the very subject of the poverty of the stimulus debate---degree 1 data alone cannot induce a preference for hierarchical structure. However, our work also supports the position of \citet{lightfoot1989child} that lower degree data is adequate for a child to learn a specific rule if they are given sufficiently rich data outside of that rule: the LM extrapolates from degree-1 QF examples if it has seen enough degree-2 declaration examples to induce a hierarchical bias. 

\paragraph{Grokking, instability, and latent structure.}
Classic grokking \citep{Power2022-hz} is different from structural grokking. While the latter entails a transition between generalization rules, the former entails a transition from memorization to generalization. Our findings clarify both scenarios. 

Structural grokking is produced by competition between linear- and hierarchical-inducing training subsets. Without competing subsets, the model immediately learns either the linear or the hierarchical rule without a gradual transition. When these rules compete, training is unstable, leaving an opening for delayed hierarchical generalization. 
Our study of data diversity has similar implications for classic grokking, where the competition is between memorized heuristics and the systematic rules required to efficiently model diverse training data. Yet again, while a strict memorization regime is relatively stable, intermediate diversity is unstable, leading to potential grokking dynamics.

In a sense, memorization is just another rule to capture the training distribution. This framework unifies the grokking literature with other phase transitions such as emergence \citep{Schaeffer2023-od} and benign interpolation \citep{Theunissen2020-pv}. Future work could develop this unified theory of the effects of data diversity and complexity.

% \clearpage
% \section*{Ethics Statement}
% This research does not present any direct ethical concerns. The work involves empirical studies of machine learning models and their behavior in language tasks. No human subjects, sensitive data, or high-stakes applications were involved in this research. Therefore, no specific ethical considerations were necessary for this work.
% \section*{Reproducibility Statement}
% All relevant details regarding the experimental setup including model architecture, hyperparameters, and data preprocessing, are included in the main text (Section~\ref{sec:model_and_training}) and appendices (Section~\ref{sec:simple_mixin}).  Additionally, the code and scripts used to run the experiments are provided in the supplementary material and will be made publicly available upon acceptance.

\section*{Limitations}

Our work leverages synthetic controlled settings that are common in the language model analysis literature. These findings, like other prominent results in grokking and training dynamics, may not generalize to larger scale or multitask settings. 
% \clearpage
\subsection*{Acknowledgements}

Tian Qin and David Alvarez-Melis were partially supported by the Kempner Institute, the Aramont Fellowship Fund, and the FAS Dean’s Competitive Fund for Promising Scholarship. We are very grateful to David Chiang, Isabel Papadimitriou, Ekdeep Singh, John Rawski, Will Merrill for their valuable feedback and discussions that helped improve this work. 

\bibliography{paperpile}

\begin{thebibliography}{55}
\providecommand{\natexlab}[1]{#1}

\bibitem[{Ahuja et~al.(2024)Ahuja, Balachandran, Panwar, He, Smith, Goyal, and Tsvetkov}]{Ahuja2024-ul}
Kabir Ahuja, Vidhisha Balachandran, Madhur Panwar, Tianxing He, Noah~A Smith, Navin Goyal, and Yulia Tsvetkov. 2024.
\newblock \href {https://arxiv.org/abs/2404.16367} {Learning syntax without planting trees: Understanding when and why transformers generalize hierarchically}.
\newblock \emph{arXiv [cs.CL]}.

\bibitem[{Barak et~al.(2022)Barak, Edelman, Goel, Kakade, Malach, and Zhang}]{Barak2022-ub}
Boaz Barak, Benjamin~L Edelman, Surbhi Goel, Sham Kakade, Eran Malach, and Cyril Zhang. 2022.
\newblock \href {https://arxiv.org/abs/2207.08799} {Hidden progress in deep learning: {SGD} learns parities near the computational limit}.
\newblock \emph{arXiv [cs.LG]}.

\bibitem[{Berlot-Attwell et~al.(2023)Berlot-Attwell, Agrawal, Carrell, Sharma, and Saphra}]{Berlot-Attwell2023-qx}
Ian Berlot-Attwell, Kumar~Krishna Agrawal, A~Michael Carrell, Yash Sharma, and Naomi Saphra. 2023.
\newblock \href {https://arxiv.org/abs/2311.08695} {Attribute diversity determines the systematicity gap in {VQA}}.
\newblock \emph{arXiv [cs.LG]}.

\bibitem[{Berwick et~al.(2011)Berwick, Pietroski, Yankama, and Chomsky}]{Berwick2011-xb}
Robert~C Berwick, Paul Pietroski, Beracah Yankama, and Noam Chomsky. 2011.
\newblock Poverty of the stimulus revisited.
\newblock \emph{Cogn. Sci.}, 35(7):1207--1242.

\bibitem[{Chen et~al.(2023)Chen, Shwartz-Ziv, Cho, Leavitt, and Saphra}]{Chen2023-fi}
Angelica Chen, Ravid Shwartz-Ziv, Kyunghyun Cho, Matthew~L Leavitt, and Naomi Saphra. 2023.
\newblock \href {https://arxiv.org/abs/2309.07311} {Sudden drops in the loss: Syntax acquisition, phase transitions, and simplicity bias in {MLMs}}.
\newblock \emph{arXiv [cs.CL]}.

\bibitem[{Chomsky(2015)}]{Chomsky2015-bg}
Noam Chomsky. 2015.
\newblock \emph{Aspects of the theory of syntax}, 50 edition.
\newblock The MIT Press. MIT Press, London, England.

\bibitem[{Choshen et~al.(2022)Choshen, Hacohen, Weinshall, and Abend}]{Choshen2022-qj}
Leshem Choshen, Guy Hacohen, Daphna Weinshall, and Omri Abend. 2022.
\newblock The grammar-learning trajectories of neural language models.
\newblock In \emph{Proceedings of the 60th Annual Meeting of the Association for Computational Linguistics (Volume 1: Long Papers)}, pages 8281--8297, Stroudsburg, PA, USA. Association for Computational Linguistics.

\bibitem[{D'Amour et~al.(2022)D'Amour, Heller, Moldovan, Adlam, Alipanahi, Beutel, Chen, Deaton, Eisenstein, Hoffman, Hormozdiari, Houlsby, Hou, Jerfel, Karthikesalingam, Lucic, Ma, McLean, Mincu, Mitani, Montanari, Nado, Natarajan, Nielson, Osborne, Raman, Ramasamy, Sayres, Schrouff, Seneviratne, Sequeira, Suresh, Veitch, Vladymyrov, Wang, Webster, Yadlowsky, Yun, Zhai, and Sculley}]{D-Amour2022-tl}
Alexander D'Amour, Katherine Heller, Dan Moldovan, Ben Adlam, Babak Alipanahi, Alex Beutel, Christina Chen, Jonathan Deaton, Jacob Eisenstein, Matthew~D Hoffman, Farhad Hormozdiari, Neil Houlsby, Shaobo Hou, Ghassen Jerfel, Alan Karthikesalingam, Mario Lucic, Yian Ma, Cory McLean, Diana Mincu, and 21 others. 2022.
\newblock Underspecification presents challenges for credibility in modern machine learning.
\newblock \emph{Journal of Machine Learning Research}, 23(226):1--61.

\bibitem[{Dodge et~al.(2020)Dodge, Ilharco, Schwartz, Farhadi, Hajishirzi, and Smith}]{Dodge2020-pb}
Jesse Dodge, Gabriel Ilharco, Roy Schwartz, Ali Farhadi, Hannaneh Hajishirzi, and Noah Smith. 2020.
\newblock \href {https://arxiv.org/abs/2002.06305} {Fine-tuning pretrained language models: Weight initializations, data orders, and early stopping}.
\newblock \emph{arXiv [cs.CL]}.

\bibitem[{Frank and Mathis(2007)}]{Frank2007-pn}
Robert Frank and Donald Mathis. 2007.
\newblock Transformational networks.
\newblock \emph{Models of Human Language Acquisition}, 22.

\bibitem[{Gao and He(2024)}]{Gao2024-fi}
Nan Gao and Qingshun He. 2024.
\newblock A dependency distance approach to the syntactic complexity variation in the connected speech of alzheimer’s disease.
\newblock \emph{Humanit. Soc. Sci. Commun.}, 11(1).

\bibitem[{Geirhos et~al.(2020)Geirhos, Jacobsen, Michaelis, Zemel, Brendel, Bethge, and Wichmann}]{Geirhos2020-ex}
Robert Geirhos, Jörn-Henrik Jacobsen, Claudio Michaelis, Richard Zemel, Wieland Brendel, Matthias Bethge, and Felix~A Wichmann. 2020.
\newblock Shortcut learning in deep neural networks.
\newblock \emph{Nat. Mach. Intell.}, 2(11):665--673.

\bibitem[{Hermann and Lampinen(2020)}]{Hermann2020-ja}
Katherine~L Hermann and Andrew~K Lampinen. 2020.
\newblock \href {https://arxiv.org/abs/2006.12433} {What shapes feature representations? exploring datasets, architectures, and training}.
\newblock \emph{arXiv [cs.LG]}.

\bibitem[{Hewitt and Manning(2019)}]{Hewitt2019-fc}
John Hewitt and Christopher~D Manning. 2019.
\newblock A structural probe for finding syntax in word representations.
\newblock In \emph{Proceedings of the 2019 Conference of the North}, pages 4129--4138, Stroudsburg, PA, USA. Association for Computational Linguistics.

\bibitem[{Hu et~al.(2023)Hu, Chen, Saphra, and Cho}]{Hu2023-vh}
Michael~Y Hu, Angelica Chen, Naomi Saphra, and Kyunghyun Cho. 2023.
\newblock \href {https://arxiv.org/abs/2308.09543} {Latent state models of training dynamics}.
\newblock \emph{arXiv [cs.LG]}.

\bibitem[{Huang et~al.(2023)Huang, Iyer, Hsu, Kumar, Chang, and Galstyan}]{Huang2023-ab}
Kuan-Hao Huang, Varun Iyer, I-Hung Hsu, Anoop Kumar, Kai-Wei Chang, and Aram Galstyan. 2023.
\newblock {ParaAMR}: A large-scale syntactically diverse paraphrase dataset by {AMR} back-translation.
\newblock In \emph{Proceedings of the 61st Annual Meeting of the Association for Computational Linguistics (Volume 1: Long Papers)}, Stroudsburg, PA, USA. Association for Computational Linguistics.

\bibitem[{Huang et~al.(2024)Huang, Hu, Han, Liu, and Sun}]{Huang2024-aw}
Yufei Huang, Shengding Hu, Xu~Han, Zhiyuan Liu, and Maosong Sun. 2024.
\newblock \href {https://arxiv.org/abs/2402.15175} {Unified view of grokking, double descent and emergent abilities: A perspective from circuits competition}.
\newblock \emph{arXiv [cs.LG]}.

\bibitem[{Juneja et~al.(2022)Juneja, Bansal, Cho, Sedoc, and Saphra}]{Juneja2022-hj}
Jeevesh Juneja, Rachit Bansal, Kyunghyun Cho, João Sedoc, and Naomi Saphra. 2022.
\newblock \href {https://arxiv.org/abs/2205.12411} {Linear connectivity reveals generalization strategies}.
\newblock \emph{arXiv [cs.LG]}.

\bibitem[{Kingma and Ba(2014)}]{Kingma2014-he}
Diederik~P Kingma and Jimmy Ba. 2014.
\newblock \href {https://arxiv.org/abs/1412.6980} {Adam: A method for stochastic optimization}.
\newblock \emph{arXiv [cs.LG]}.

\bibitem[{Lakshminarayanan et~al.(2016)Lakshminarayanan, Pritzel, and Blundell}]{Lakshminarayanan2016-fb}
Balaji Lakshminarayanan, Alexander Pritzel, and Charles Blundell. 2016.
\newblock \href {https://arxiv.org/abs/1612.01474} {Simple and scalable predictive uncertainty estimation using deep ensembles}.
\newblock \emph{arXiv [stat.ML]}.

\bibitem[{Lightfoot(1989)}]{lightfoot1989child}
David Lightfoot. 1989.
\newblock The child's trigger experience: Degree-0 learnability.
\newblock \emph{Behavioral and brain sciences}, 12(2):321--334.

\bibitem[{Linzen et~al.(2016)Linzen, Dupoux, and Goldberg}]{Linzen2016-vx}
Tal Linzen, Emmanuel Dupoux, and Yoav Goldberg. 2016.
\newblock Assessing the ability of {LSTMs} to learn syntax-sensitive dependencies.
\newblock \emph{Trans. Assoc. Comput. Linguist.}, 4:521--535.

\bibitem[{Liu et~al.(2022)Liu, Michaud, and Tegmark}]{Liu2022-mj}
Ziming Liu, Eric~J Michaud, and Max Tegmark. 2022.
\newblock \href {https://arxiv.org/abs/2210.01117} {Omnigrok: Grokking beyond algorithmic data}.
\newblock \emph{arXiv [cs.LG]}.

\bibitem[{Lubana et~al.(2024)Lubana, Kawaguchi, Dick, and Tanaka}]{Lubana2024-ed}
Ekdeep~Singh Lubana, Kyogo Kawaguchi, Robert~P Dick, and Hidenori Tanaka. 2024.
\newblock \href {https://arxiv.org/abs/2408.12578} {A percolation model of emergence: Analyzing transformers trained on a formal language}.
\newblock \emph{arXiv [cs.LG]}.

\bibitem[{MacWhinney(2014)}]{MacWhinney2014-jq}
Brian MacWhinney. 2014.
\newblock \emph{The childes project: Tools for analyzing talk, volume {II}: The database}, 3 edition.
\newblock Psychology Press, London, England.

\bibitem[{McCoy et~al.(2018)McCoy, Frank, and Linzen}]{McCoy2018-uv}
R~Thomas McCoy, Robert Frank, and Tal Linzen. 2018.
\newblock \href {https://arxiv.org/abs/1802.09091} {Revisiting the poverty of the stimulus: hierarchical generalization without a hierarchical bias in recurrent neural networks}.
\newblock \emph{arXiv [cs.CL]}.

\bibitem[{McCoy et~al.(2020{\natexlab{a}})McCoy, Frank, and Linzen}]{McCoy2020-pj}
R~Thomas McCoy, Robert Frank, and Tal Linzen. 2020{\natexlab{a}}.
\newblock Does syntax need to grow on trees? sources of hierarchical inductive bias in sequence-to-sequence networks.
\newblock \emph{Trans. Assoc. Comput. Linguist.}, 8:125--140.

\bibitem[{McCoy et~al.(2019)McCoy, Pavlick, and Linzen}]{McCoy2019-br}
R~Thomas McCoy, Ellie Pavlick, and Tal Linzen. 2019.
\newblock \href {https://arxiv.org/abs/1902.01007} {Right for the wrong reasons: Diagnosing syntactic heuristics in natural language inference}.
\newblock \emph{arXiv [cs.CL]}.

\bibitem[{McCoy et~al.(2020{\natexlab{b}})McCoy, Frank, and Linzen}]{McCoyUnknown-uy}
Tom McCoy, Robert Frank, and Tal Linzen. 2020{\natexlab{b}}.
\newblock Does syntax need to grow on trees? sources of hierarchical inductive bias in sequence-to-sequence networks.
\newblock \url{https://rtmccoy.com/rnn\_hierarchical\_biases.html}.
\newblock Accessed: 2024-11-20.

\bibitem[{Merrill et~al.(2023)Merrill, Tsilivis, and Shukla}]{Merrill2023-an}
William Merrill, Nikolaos Tsilivis, and Aman Shukla. 2023.
\newblock \href {https://arxiv.org/abs/2303.11873} {A tale of two circuits: Grokking as competition of sparse and dense subnetworks}.
\newblock \emph{arXiv [cs.LG]}.

\bibitem[{Mueller et~al.(2022)Mueller, Frank, Linzen, Wang, and Schuster}]{Mueller2022-rm}
Aaron Mueller, Robert Frank, Tal Linzen, Luheng Wang, and Sebastian Schuster. 2022.
\newblock Coloring the blank slate: Pre-training imparts a hierarchical inductive bias to sequence-to-sequence models.
\newblock In \emph{Findings of the Association for Computational Linguistics: ACL 2022}, Stroudsburg, PA, USA. Association for Computational Linguistics.

\bibitem[{Mueller and Linzen(2023)}]{Mueller2023-xq}
Aaron Mueller and Tal Linzen. 2023.
\newblock \href {https://arxiv.org/abs/2305.19905} {How to plant trees in language models: Data and architectural effects on the emergence of syntactic inductive biases}.
\newblock \emph{arXiv [cs.CL]}.

\bibitem[{Mueller et~al.(2024)Mueller, Webson, Petty, and Linzen}]{Mueller2024-qz}
Aaron Mueller, Albert Webson, Jackson Petty, and Tal Linzen. 2024.
\newblock In-context learning generalizes, but not always robustly: The case of syntax.
\newblock In \emph{Proceedings of the 2024 Conference of the North American Chapter of the Association for Computational Linguistics: Human Language Technologies (Volume 1: Long Papers)}, pages 4761--4779, Stroudsburg, PA, USA. Association for Computational Linguistics.

\bibitem[{Murty et~al.(2023)Murty, Sharma, Andreas, and Manning}]{Murty2023-xp}
Shikhar Murty, Pratyusha Sharma, Jacob Andreas, and Christopher Manning. 2023.
\newblock Grokking of hierarchical structure in vanilla transformers.
\newblock In \emph{Proceedings of the 61st Annual Meeting of the Association for Computational Linguistics (Volume 2: Short Papers)}, Stroudsburg, PA, USA. Association for Computational Linguistics.

\bibitem[{Murty et~al.(2022)Murty, Sharma, Andreas, and Manning}]{Murty2022-lw}
Shikhar Murty, Pratyusha Sharma, Jacob Andreas, and Christopher~D Manning. 2022.
\newblock \href {https://arxiv.org/abs/2211.01288} {Characterizing intrinsic compositionality in transformers with tree projections}.
\newblock \emph{arXiv [cs.CL]}.

\bibitem[{Naik et~al.(2018)Naik, Ravichander, Sadeh, Rose, and Neubig}]{Naik2018-og}
Aakanksha Naik, Abhilasha Ravichander, Norman Sadeh, Carolyn Rose, and Graham Neubig. 2018.
\newblock Stress test evaluation for natural language inference.
\newblock In \emph{Proceedings of the 27th International Conference on Computational Linguistics}, pages 2340--2353.

\bibitem[{Nanda et~al.(2023)Nanda, Chan, Lieberum, Smith, and Steinhardt}]{Nanda2023-zm}
Neel Nanda, Lawrence Chan, Tom Lieberum, Jess Smith, and Jacob Steinhardt. 2023.
\newblock \href {https://arxiv.org/abs/2301.05217} {Progress measures for grokking via mechanistic interpretability}.
\newblock \emph{arXiv [cs.LG]}.

\bibitem[{Olsson et~al.(2022)Olsson, Elhage, Nanda, Joseph, DasSarma, Henighan, Mann, Askell, Bai, Chen, Conerly, Drain, Ganguli, Hatfield-Dodds, Hernandez, Johnston, Jones, Kernion, Lovitt, Ndousse, Amodei, Brown, Clark, Kaplan, McCandlish, and Olah}]{Olsson2022-ed}
Catherine Olsson, Nelson Elhage, Neel Nanda, Nicholas Joseph, Nova DasSarma, Tom Henighan, Ben Mann, Amanda Askell, Yuntao Bai, Anna Chen, Tom Conerly, Dawn Drain, Deep Ganguli, Zac Hatfield-Dodds, Danny Hernandez, Scott Johnston, Andy Jones, Jackson Kernion, Liane Lovitt, and 7 others. 2022.
\newblock \href {https://arxiv.org/abs/2209.11895} {In-context learning and induction heads}.
\newblock \emph{arXiv [cs.LG]}.

\bibitem[{Papadimitriou and Jurafsky(2020)}]{Papadimitriou2020-la}
Isabel Papadimitriou and Dan Jurafsky. 2020.
\newblock Learning music helps you read: Using transfer to study linguistic structure in language models.
\newblock In \emph{Proceedings of the 2020 Conference on Empirical Methods in Natural Language Processing (EMNLP)}, Stroudsburg, PA, USA. Association for Computational Linguistics.

\bibitem[{Papadimitriou and Jurafsky(2023)}]{Papadimitriou2023-gj}
Isabel Papadimitriou and Dan Jurafsky. 2023.
\newblock \href {https://arxiv.org/abs/2304.13060} {Injecting structural hints: Using language models to study inductive biases in language learning}.
\newblock \emph{arXiv [cs.CL]}.

\bibitem[{Park et~al.(2024)Park, Lubana, Pres, and Tanaka}]{Park2024-ri}
Core~Francisco Park, Ekdeep~Singh Lubana, Itamar Pres, and Hidenori Tanaka. 2024.
\newblock \href {https://arxiv.org/abs/2412.01003} {Competition dynamics shape algorithmic phases of in-context learning}.
\newblock \emph{arXiv [cs.LG]}.

\bibitem[{Petty and Frank(2021)}]{Petty2021-pe}
Jackson Petty and Robert Frank. 2021.
\newblock \href {https://arxiv.org/abs/2109.12036} {Transformers generalize linearly}.
\newblock \emph{arXiv [cs.CL]}.

\bibitem[{Power et~al.(2022)Power, Burda, Edwards, Babuschkin, and Misra}]{Power2022-hz}
Alethea Power, Yuri Burda, Harri Edwards, Igor Babuschkin, and Vedant Misra. 2022.
\newblock \href {https://arxiv.org/abs/2201.02177} {Grokking: Generalization beyond overfitting on small algorithmic datasets}.
\newblock \emph{arXiv [cs.LG]}.

\bibitem[{Ramírez et~al.(2022)Ramírez, Baez, Berro, Benatallah, and Casati}]{Ramirez2022-mx}
Jorge Ramírez, Marcos Baez, Auday Berro, Boualem Benatallah, and Fabio Casati. 2022.
\newblock Crowdsourcing syntactically diverse paraphrases with diversity-aware prompts and workflows.
\newblock In \emph{Advanced Information Systems Engineering: 34th International Conference, CAiSE 2022, Leuven, Belgium, June 6–10, 2022, Proceedings}, page 253–269, Berlin, Heidelberg. Springer-Verlag.

\bibitem[{Saphra and Lopez(2018)}]{Saphra2018-xx}
Naomi Saphra and Adam Lopez. 2018.
\newblock \href {https://arxiv.org/abs/1811.00225} {Understanding learning dynamics of language models with {SVCCA}}.
\newblock \emph{arXiv [cs.CL]}.

\bibitem[{Saphra and Lopez(2019)}]{Saphra2019-sq}
Naomi Saphra and Adam Lopez. 2019.
\newblock Understanding learning dynamics of language models with.
\newblock In \emph{Proceedings of the 2019 Conference of the North}, pages 3257--3267, Stroudsburg, PA, USA. Association for Computational Linguistics.

\bibitem[{Schaeffer et~al.(2023)Schaeffer, Miranda, and Koyejo}]{Schaeffer2023-od}
Rylan Schaeffer, Brando Miranda, and Sanmi Koyejo. 2023.
\newblock \href {https://arxiv.org/abs/2304.15004} {Are emergent abilities of large language models a mirage?}
\newblock \emph{arXiv [cs.AI]}.

\bibitem[{Sellam et~al.(2021)Sellam, Yadlowsky, Tenney, Wei, Saphra, D'Amour, Linzen, Bastings, Turc, Eisenstein, Das, and Pavlick}]{Sellam2021-rz}
Thibault Sellam, Steve Yadlowsky, Ian Tenney, Jason Wei, Naomi Saphra, Alexander D'Amour, Tal Linzen, Jasmijn Bastings, Iulia~Raluca Turc, Jacob Eisenstein, Dipanjan Das, and Ellie Pavlick. 2021.
\newblock The {MultiBERTs}: {BERT} reproductions for robustness analysis.
\newblock In \emph{International Conference on Learning Representations}.

\bibitem[{Song et~al.(2024)Song, Lothritz, Tang, Bissyandé, and Klein}]{Song2024-cg}
Yewei Song, Cedric Lothritz, Daniel Tang, Tegawendé~F Bissyandé, and Jacques Klein. 2024.
\newblock \href {https://arxiv.org/abs/2404.08817} {Revisiting code similarity evaluation with abstract syntax tree edit distance}.
\newblock \emph{arXiv [cs.CL]}.

\bibitem[{Theunissen et~al.(2020)Theunissen, Davel, and Barnard}]{Theunissen2020-pv}
Marthinus~Wilhelmus Theunissen, Marelie Davel, and Etienne Barnard. 2020.
\newblock Benign interpolation of noise in deep learning.
\newblock \emph{S. Afr. Comput. J.}, 32(2).

\bibitem[{Varma et~al.(2023)Varma, Shah, Kenton, Kramár, and Kumar}]{Varma2023-iq}
Vikrant Varma, Rohin Shah, Zachary Kenton, János Kramár, and Ramana Kumar. 2023.
\newblock \href {https://arxiv.org/abs/2309.02390} {Explaining grokking through circuit efficiency}.
\newblock \emph{arXiv [cs.LG]}.

\bibitem[{Wexler(1980)}]{wexler1980formal}
Kenneth Wexler. 1980.
\newblock Formal principles of language acquisition.

\bibitem[{Zhang and Shasha(1989)}]{Zhang1989-ma}
Kaizhong Zhang and Dennis Shasha. 1989.
\newblock Simple fast algorithms for the editing distance between trees and related problems.
\newblock \emph{SIAM J. Comput.}, 18(6):1245--1262.

\bibitem[{Zhou et~al.(2020)Zhou, Nie, Tan, and Bansal}]{Zhou2020-xt}
Xiang Zhou, Yixin Nie, Hao Tan, and Mohit Bansal. 2020.
\newblock The curse of performance instability in analysis datasets: Consequences, source, and suggestions.
\newblock In \emph{Proceedings of the 2020 Conference on Empirical Methods in Natural Language Processing}.

\bibitem[{Zhu et~al.(2024)Zhu, Fu, Zhou, and Lin}]{Zhu2024-nz}
Xuekai Zhu, Yao Fu, Bowen Zhou, and Zhouhan Lin. 2024.
\newblock \href {https://arxiv.org/abs/2401.10463} {Critical data size of language models from a grokking perspective}.
\newblock \emph{arXiv [cs.CL]}.

\end{thebibliography}
% \bibliographystyle{acl_natbib}

% \bibliography{paperpile,iclr2025_conference}
% \bibliographystyle{iclr2025_conference}

\clearpage

\appendix

\section{Related Work Extended}
\label{appdx:related}
\subsection{Syntax and Hierarchical Generalization}
 While works mentioned in Section~\ref{sec:syntax_related} focused on models trained from scratch, another line of research examined the inductive bias of pretrained models. \citet{Mueller2024-qz, Mueller2023-xq} pretrained transformers on text corpora such as Wikipedia and CHILDES \citep{MacWhinney2014-jq} before fine-tuning them on the question formation task. They found that exposure to large amounts of natural language data enables transformers to generalize hierarchically.

Instead of using the question formation task as a probe, \citet{Hewitt2019-fc, Murty2022-lw} directly interpreted model's internal representation to understand whether transformers constrain their computations to to follow tree-structure patterns. \citet{Hewitt2019-fc} demonstrated that the syntax tress are embedded in model's representation space. Similarly, \citet{Murty2022-lw} projects transformers into a tree-structured network, and showed that transformers become more tree-like over the course of training on language data. 

\citet{Papadimitriou2023-gj, Papadimitriou2020-la} and \citet{Mueller2022-rm} also studied how pretraining data could introduce an inductive bias in language acquisition. \citet{Papadimitriou2023-gj} specifically identified that by pretraining models on data with a recursive structure  their performance when later finetuning them on natural language. This finding is closely related to our conclusions around the importance of recursive center embeddings.

\begin{figure*}[t!]
    \centering
    \includegraphics[width=0.49\linewidth]{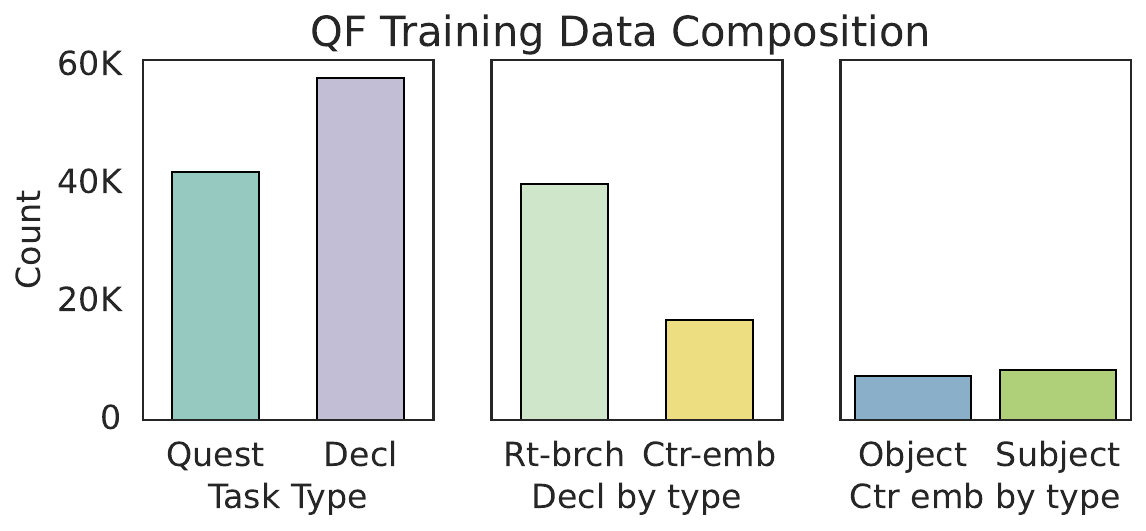}
    \includegraphics[width=0.49\linewidth]{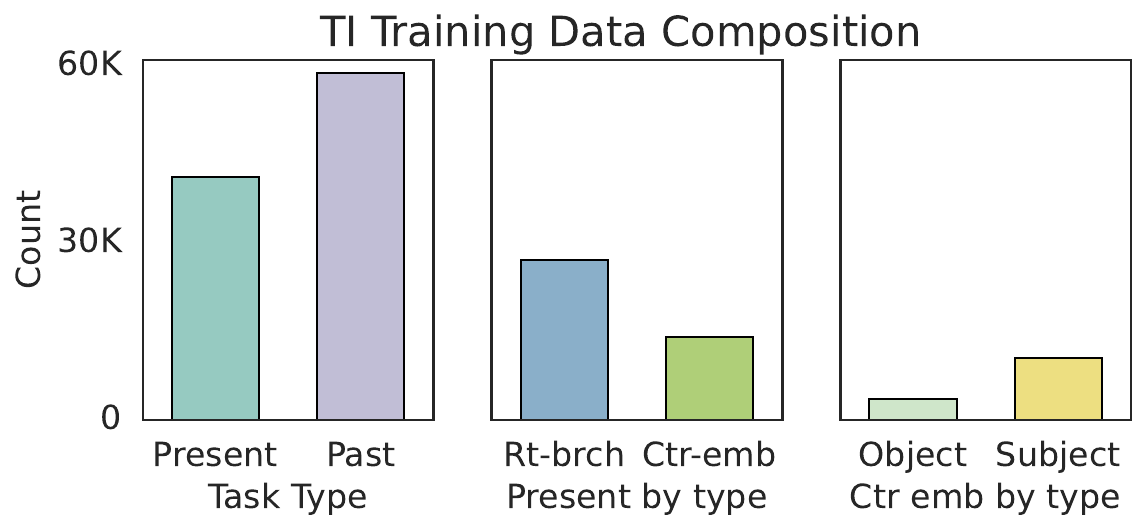}
    \caption{\textbf{Components of the original QF and TI training data.} 
    \textit{Left:} QF training data contains samples of two tasks types: question formation and declaration copying. We break down samples in the declaration copying task by branching type. We also breakdown center-embedded sentences based on whether the main subject serves the subject or object in the embedded clause.
    \textit{Right:} TI training data also contains samples of two task types: tense inflection and past tense copying. Similar to QF, we breakdown tense inflection samples by branching types, and breakdown center-embedded sentences in the tense inflection samples by subject/object type. 
    }
    \label{fig:data_detail}
\end{figure*}
% \paragraph{Random variation} 
\subsection{Random Variation}
Specific training choices, such as hyperparameters, are crucial to model outcomes. However, even when controlling for these factors, training machine learning models remains inherently stochastic—models can be sensitive to random initialization and the order of training examples. \citet{Zhou2020-xt, D-Amour2022-tl, Naik2018-og} reported significant performance differences across model checkpoints on various analysis and stress test sets. \citet{Zhou2020-xt} further found that instability extends throughout the training curve, not just in final outcomes. To investigate the source of this inconsistency, \citet{Dodge2020-pb} compared the effects of weight initialization and data order, concluding that both factors contribute equally to variations in out-of-sample performance.

Similarly, \citet{Sellam2021-rz} found that repeating the pre-training process on BERT models can result in significantly different performances on downstream tasks. To promote more robust experimental testing, they introduced a set of 25 BERT-BASE checkpoints to ensure that experimental conclusions are not influenced by artifacts, such as specific instances of the model. In this work, we also observe training inconsistencies across runs on OOD data, both during training and at convergence. Unlike prior studies that focus on implications of random variations on experimental design, we study the source of training inconsistencies and link these inconsistencies to simplicity bias and the characteristics of the training data.

% \paragraph{Simplicity bias} 
\subsection{Simplicity Bias}
Models often favor simpler functions early in training, a phenomenon known as simplicity bias \citep{Hermann2020-ja}, which is also common in LMs. \citet{Choshen2022-qj} found that early LMs behave like n-gram models, and \citet{Saphra2019-sq} observed that early LMs learn simplified versions of the language modeling task. \citet{McCoy2019-br} showed that even fully trained models can rely on simple heuristics, like lexical overlap, to perform well on Natural Language Inference (NLI) tasks. \citet{Chen2023-fi} further explored the connection between training dynamics and simplicity bias, showing that simpler functions learned early on can continue to influence fully trained models, and mitigating this bias can have long-term effects on training outcomes.

Phase transitions have been identified as markers of shifts from simplistic heuristics to more complex model behavior, often triggered by the amount of training data or model size. In language models, \citet{Olsson2022-ed} showed that the emergence of induction heads in autoregressive models is linked to handling longer context sizes and in-context learning. Similar phase transitions have been studied in non-language domains, such as algorithmic tasks \citep{Power2022-hz, Merrill2023-an} and arithmetic tasks \citep{Nanda2023-zm, Barak2022-ub}.

In the context of hierarchical generalization, \citet{Ahuja2024-ul} used a Bayesian approach to analyze the simplicity of hierarchical versus linear rules in modeling English syntax. They argued that transformers favor the hierarchical rule because it is simpler than the linear rule. However, their model fails to explain (1) why learning the hierarchical rule is delayed (i.e., after learning the linear rule) and (2) why hierarchical generalization is inconsistent across runs. In this work, we offer a different perspective, showing that a model's simplicity bias towards either rule is driven by the characteristics of the training data.

\section{Training Data Samples}
\label{appdx:data_sample}

\begin{figure*}[ht]
    \centering
    \includegraphics[width=0.6\linewidth]{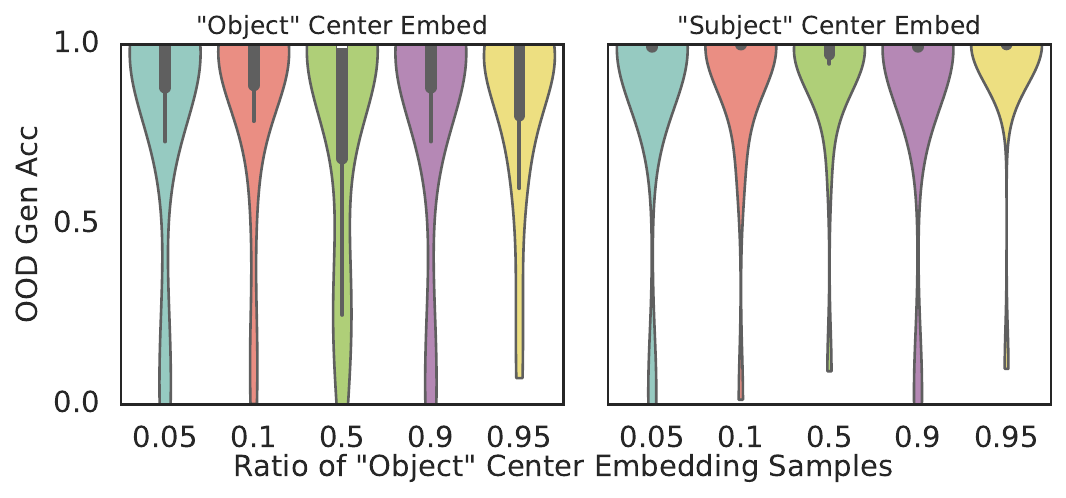}
    \caption{\textbf{Both subtypes of center-embedded sentences induce hierarchical generalization in QF.} We train models on datasets containing different ratios of object-type v.s. subject-type center-embedded sentences. We then evaluate on models on two OOD generalization set, one containing unambiguous object-type center-embedded sentences and the other unambiguous subject-type center-embedded sentences.}
    \label{fig:qf_rc_type}
\end{figure*}
\begin{figure*}[ht]
    \centering
    \includegraphics[width=0.9\linewidth]{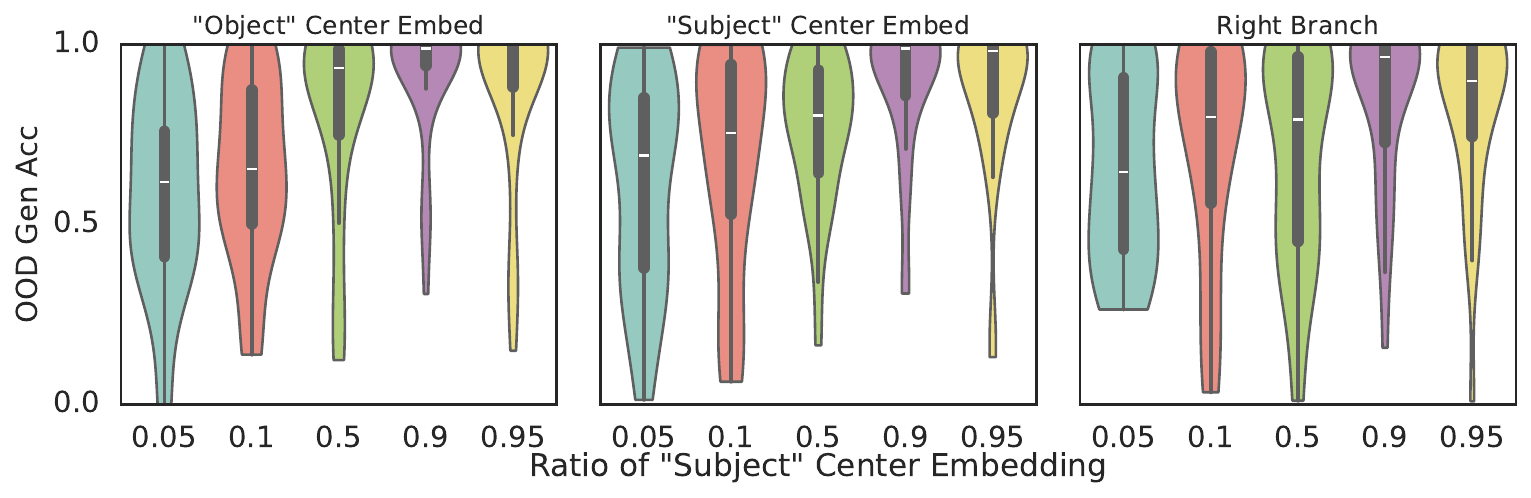}
    \caption{\textbf{Subject-type center-embedded sentences gives a stronger bias towards hierarchical generalization in TI.} We train models on datasets containing different ratios of object-type v.s. subject-type center-embedded sentences. We then evaluate on models on three OOD generalization set, one containing unambiguous object-type center-embedded sentences, one unambiguous subject-type center-embedded sentences, and one unambiguous right-branching sentences.}
    \label{fig:ti_rc_type}
\end{figure*}

\subsection{Question Formation}

We use the term \textbf{declarations} to refer to the declaration copying task and \textbf{questions} refer to the question formation task. Here are two examples randomly taken from the training data: 

\begin{itemize}[itemsep=2pt,labelindent=0pt,topsep=0pt,parsep=0pt,partopsep=1pt, align=left, leftmargin=*]
    \item Declaration Example: \texttt{our zebra doesn't applaud the unicorn . decl	our zebra doesn't applaud the unicorn .}
    \item Question Example: \texttt{some unicorns do move . quest	do some unicorns move ?}
\end{itemize}

Both tasks begin with an input declarative sentence, followed by a task indicator token (\texttt{decl} or \texttt{quest}), and end with the output. During training, the entire sequence is used in the causal language modeling objective. The ID validation set and the OOD generalization set only contain question formation samples. In Figure~\ref{fig:data_detail} (\textit{left}), we show a breakdown of two task types in QF training data.

\subsection{Tense Inflection}
\begin{itemize}[itemsep=2pt,labelindent=0pt,topsep=0pt,parsep=0pt,partopsep=1pt, align=left, leftmargin=*]
   \item  Past Example: \texttt{our peacocks above our walruses amused your zebras . PAST	our peacocks above our walruses amused your zebras .}
    \item Present Example: \texttt{your unicorns that our xylophones comforted swam . PRESENT	your unicorns that our xylophones comfort swim .}
\end{itemize}
The tense inflection task is indicate by the \texttt{PRESENT} token. The secondary task only requires repeating the given sentence, which is always in the past tense, and the copying task is marked by the \texttt{PAST} token. In Appendix~\ref{appdx:ti_secondary}, show that the past-tense copying task is not necessary.

\section{Further Partitions on Center-Embedded Sentences}
\label{appdx:obj_sbj_ctr_breakdown}

\subsection{Two Subtypes of Center-Embedded Sentences}
In Section~\ref{sec:data_complexity}, we showed that center-embedded sentences drive hierarchical generalization in both the QF and TI tasks. Here, we further partition center-embedded sentences based on the syntactic role of the \textit{main subject} (i.e., the subject of the main clause) within the modifying clause. Specifically, we classify them into two types:

\begin{enumerate}[itemsep=4pt,labelindent=15pt,topsep=0pt,parsep=0pt,partopsep=1pt,align=left,leftmargin=*]
    \item \textbf{Subject-type}: The main subject serves as the \textbf{subject} within the clause.

    Example: \textit{The keys that unlock the cabinet are on the table.}

    \item \textbf{Object-type}: The main subject serves as the \textbf{object} within the clause.
    
    Example:\textit{ The keys that the bear uses are on the table.}
\end{enumerate}

This partition is motivated by their distinct subject-verb dependency patterns. In subject-type sentences, both the main verb (from the main clause) and the embedded verb (from the relative clause) depend on the main subject. In contrast, object-type sentences exhibit a nested subject-verb structure. Our goal is to investigate whether differences in subject-verb dependency patterns influence the model’s preference for the hierarchical rule.

\subsection{QF Task Results}
The original QF training data contains roughly equal amount of two subtypes of center-embedded declarations, shown in Figure~\ref{fig:data_detail} (\textit{left}). We investigate whether the two subtypes of center-embedded sentences differentially influence the model's preference for the hierarchical rule in the QF task. For all training data variants, we fix 50K question formation samples and 50K declaration copying samples, with the latter containing only center-embedded sentences but varying the ratio between the two subtypes. To analyze generalization behavior on a more granular level, we partition the generalization set (composed solely of center-embedded sentences) into the two subtypes as well. Models are trained on 30 random seeds, and results are shown in Figure~\ref{fig:qf_rc_type}. Regardless of the data mix, the model consistently favors the hierarchical rule across both partitions of the generalization set. This suggests that, for question formation, both subtypes of center-embedded sentences equally contribute to the model's ability to identify the main auxiliary.

\subsection{TI Task Results} 
The original TI training data contains almost twice amount of subject-type center-embedded sentences than object-type ones, shown in Figure~\ref{fig:data_detail} (\textit{left}).
We repeat a similar experiment for the TI task, fixing the total number of tense inflection samples to 100K. As shown in Section~\ref{sec:ti_result}, models exhibit the strongest hierarchical generalization when trained on primarily center-embedded sentences. Therefore, in the following data variants, 99\% of the samples are center-embedded sentences, with the remaining 1\% being right-branching sentences. Within the center-embedded samples, we vary the ratio between the two subtypes. To evaluate generalization, we split the generalization set into three groups: the two subtypes of center-embedded sentences and right-branching sentences. Models trained on 30 random seeds show that, across all three generalization sets, accuracy is positively correlated with the proportion of subject-type center-embedded sentences (Figure~\ref{fig:ti_rc_type}). However, even when models are trained predominantly on object-type center-embedded sentences (teal violins in Figure~\ref{fig:ti_rc_type}), they still show a clear preference toward hierarchical generalization. Thus, while both subtypes drive hierarchical generalization in TI, subject-type center-embedded sentences have a stronger effect.

\section{Varying Data Ratios for Question Formation}
\label{sec:simple_mixin}

\begin{figure}[h]
    \centering
    % \begin{figure}
        \centering
        \includegraphics[width=0.7\linewidth]{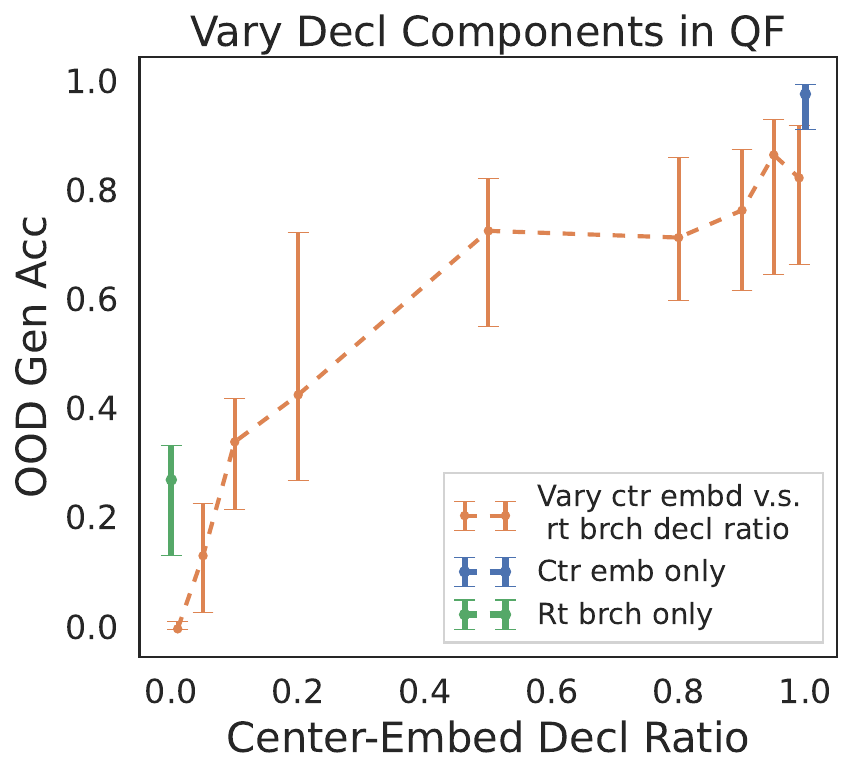}
        \caption{Hierarchical generalization in QF is sensitive to compositions of declaration-copying samples.} 
        \label{fig:simplicity_contamination}
\end{figure}

\paragraph{Data composition details} We construct variations of the training data using the following procedure. Each new dataset contains 50K questions (reused from the original data) and 50K declarations, where we control the ratio between center-embedded and right-branching sentences. These datasets are used for the experiments in Section~\ref{sec:intra_inter}. To generate additional declarations, we keep the distribution of the unique syntax structures in original dataset. Specifically, for each sentence in the original data, we extract the syntax tree using the CGF rules and resample words from the vocabulary to create new sentence samples. 

\paragraph{Sensitivity to data compositions}  We use the five datasets above to examine how different mix ratios affect a model's preference towards the hierarchical generalization. The median generalization accuracy, along with error bars representing the 35th and 65th percentiles, is shown in Figure~\ref{fig:simplicity_contamination}. First, note that there is a sharp performance drop between the blue bar and the right-most orange bar. This sharp transition indicates that mixing in as little as 1\% of right-branching declarations significantly reduces the model's likelihood of generalizing hierarchically. Interestingly, when the dataset is predominantly right-branching declarations, models consistently achieve 0\% generalization accuracy, indicating a strong preference for the linear rule across all training runs. However, note that there is another sharp transition between the green bar and the left-most orange bar. This transition indicates that as soon as we remove the 1\% of center-embedded sentences, the model fails to learn either the linear rule or the hierarchical rule. As a result, the generalization accuracy is close to random guess ($\sim 25\%$). This transition is closely studied in Section~\ref{sec:data_diveristy}, where we examine how data diversity leads to rule commitment. 

\begin{figure}[h]   
        \centering
        \includegraphics[width=0.73\linewidth]{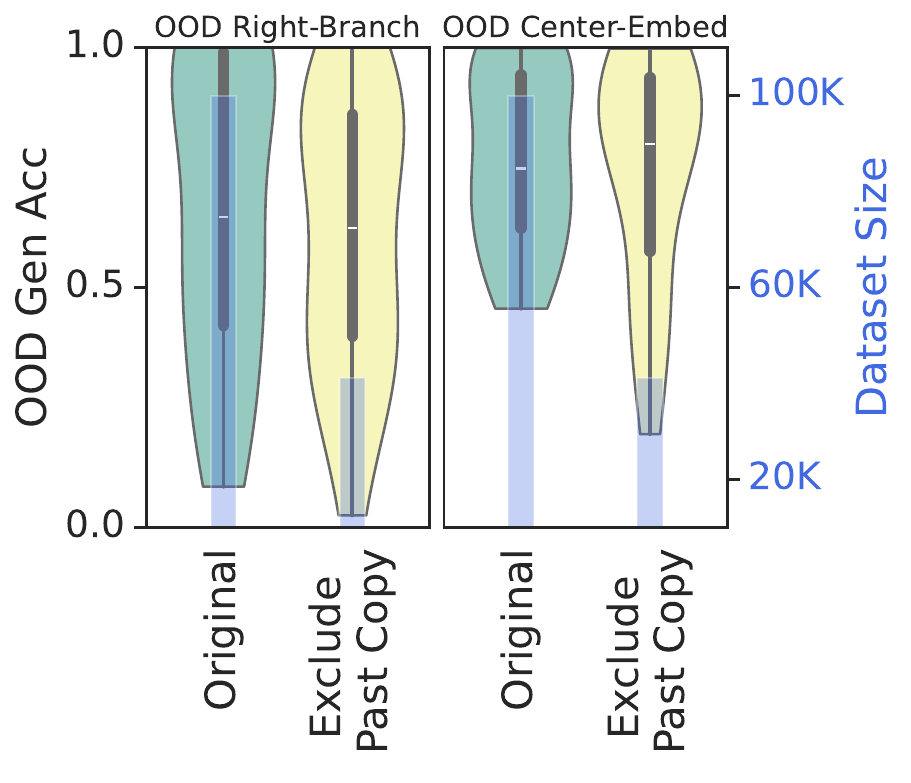}
        \caption{Past-copy task is not necessary to induce hierarchical generalization in TI.
        }
        \label{fig:tense_original_violin}
\end{figure}

\begin{figure}[h]
    \centering
    \includegraphics[width=0.7\linewidth]{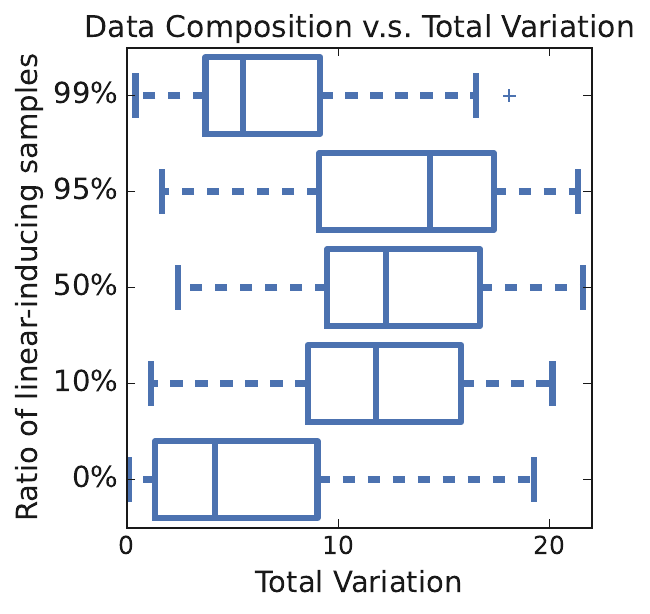}
    \caption{\textbf{Training is unstable when different subsets of data compete.} Balanced mixtures of right-branching and center-embedded sentences have higher total variation than mixtures dominated by one or the other subset.}
    \label{fig:data_drives_inconsistency}
\end{figure}

\textbf{Additional Analysis for Section~\ref{sec:intra_inter}}
Figure~\ref{fig:data_drives_inconsistency} shows the relationship between data homogeneity and training stability. When the training data is dominated by either linearity-inducing  (99\% linear) or hierarchy-inducing  (0\% linear) examples, more random seeds lead to stable OOD curves. When the training data is a heterogeneous mix instead, potential rules compete, leading to a higher proportion of unstable training runs.

\section{Additional Results on Tense Inflection} 
\subsection{A Secondary Task is Not Necessary}
\label{appdx:ti_secondary}

In the original of TI training data \citep{McCoy2020-pj}, a secondary task is also included to mimic the question formation training data. In this secondary task, instead of transforming a sentence from the past tense to the present tense, the model simply needs to repeat it. For concrete examples, see Appendix~\ref{appdx:data_sample}. Figure~\ref{fig:data_detail} (\textit{right}) shows a breakdown of the two tasks in the original TI training data. In experiments conducted in Section~\ref{sec:ti_task}, we have eliminated the used of this secondary task because center-embedded sentences can be included in the tense inflection training samples \textit{without} violating the ambiguity requirement. Here, we use the training data originally proposed by \citet{McCoy2020-pj} to confirm that the use of secondary task is indeed not necessary. Specifically, we remove all the past-tense-copying samples from the original training data and train models on the tense-inflection task only. We evaluate the model's generalization performance on two OOD set containing unambiguous right-branching and unambiguous center-embedded sentences, shown in Figure~\ref{fig:tense_original_violin}. We can see that the model's OOD performances are the same with or without the secondary task.

\subsection{Training Instability and Rule Commitment for TI}
\label{appdx:tense_tv}
\begin{figure*}[h]
    \centering
    \includegraphics[width=1.0\linewidth]{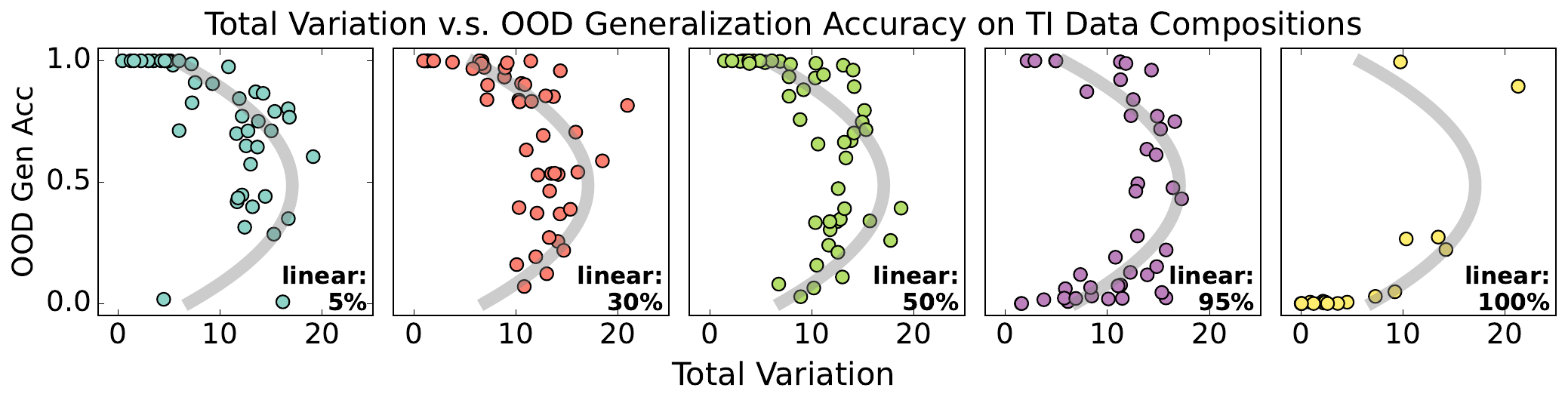}
    \caption{\textbf{Total Variation v.s. final generalization accuracy for TI task.} Similar to Figure~\ref{fig:intra_inter_variance}, we observe the same horseshoe shaped behavior between training stability and final generalization accuracy on right-branching sentences for the TI task.
    }
    \label{fig:intra_inter_variance_tense}
\end{figure*}

We repeat the same total variation analysis in Section~\ref{sec:stability} for the tense inflection task. We use the data mixes from Section~\ref{sec:ti_result}. Specifically, we include only tense inflection samples and vary the ratio between linearity-inducing (i.e., right-branching) and hierarchy-inducing (i.e., center-embedded) sentences. In Section~\ref{sec:ti_result}, we have already concluded that the hierarchical rule is \textit{always} preferred for center-embedded sentences regardless of data mixes. For this reason, we are interested in examining the rule preference and training stability for unambiguous right-branching sentences.
In Figure~\ref{fig:intra_inter_variance_tense} we visualize the relationship between total variation and the final generalization accuracy on unambiguous right-branching sentences. The qualitative behavior is similar to what we have observed in QF (Section~\ref{sec:intra_inter}).

\section{Training Instability}
\label{appdx:training_instabilty}

\begin{figure*}[h]
    \centering
    \includegraphics[width=.7\textwidth]{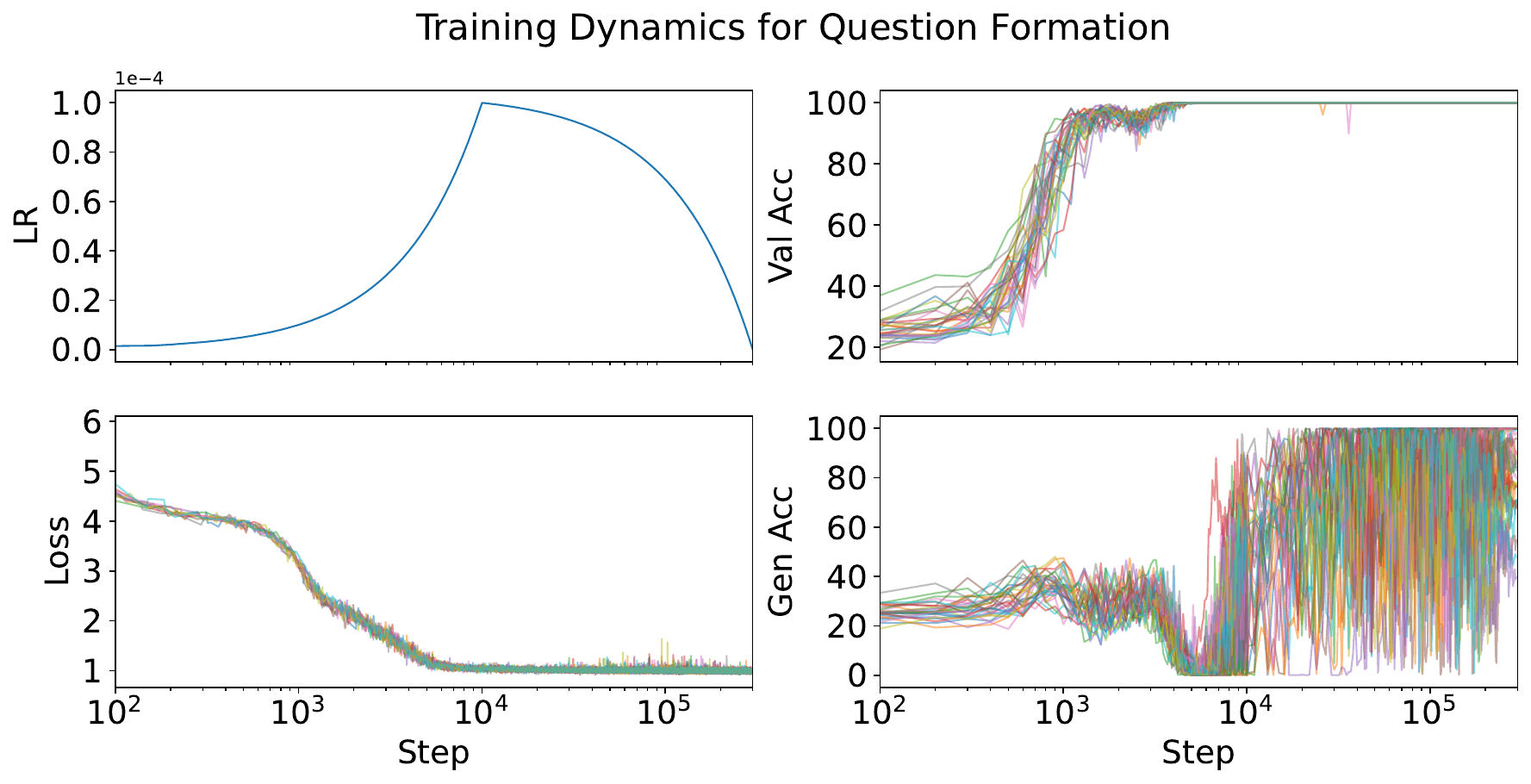}
    \caption{\textbf{Training dynamics on original QF data across 50 random seeds.} Training loss (\textit{lower left}) and in-distribution validation accuracy  (\textit{top right}) is stable during training and consistent across random seeds. In contrast, the model's performance on OOD generalization set (\textit{lower right}) is both unstable during training and inconsistent across seeds. The instability and inconsistency is most prominent during grokking (i.e., when training loss has converged). Even with a learning rate decay (\textit{top right}), the OOD behaviors for some seeds remain unstable throughout training.
    }
    \label{fig:grokking_inconsistency}
% \end{wrapfigure}
\end{figure*}
In Figure~\ref{fig:grokking_inconsistency}, we visualize the training dynamics for 30 independent runs when trained on the original QF data. Each run differs in both model initialization and data order. Notice that the training dynamics for runs exhibit grokking behaviors: OOD generalization is delayed when compared to training loss convergence and validation performance convergence. These runs share a similar progression in training loss, validation accuracy, and generalization accuracy up until moment when the training loss converges. 
Interestingly, after convergence on training loss, all runs reach $0\%$ on the generalization set, indicating that the model strictly prefers linear rules on OOD data. 
After that, models start to achieve non-trivial performance in generalization accuracy.
However, for many runs the generalization accuracy does not increase monotonically. Instead, we observe massive swings in generalization accuracy during this training period as well as large inconsistency across different seeds.
Overall, training is \textit{always} stable for ID data while the performance for OOD data is inconsistent across seeds. 
We visualize runs with different of total variation values in Figure~\ref{fig:sample_runs}.

\section{Data Diversity and Memorization Patterns}
\subsection{Memorization Patterns}
\label{appdx:memorizaition}
\begin{figure*}[h]
    \centering
    \includegraphics[width=0.8\linewidth]{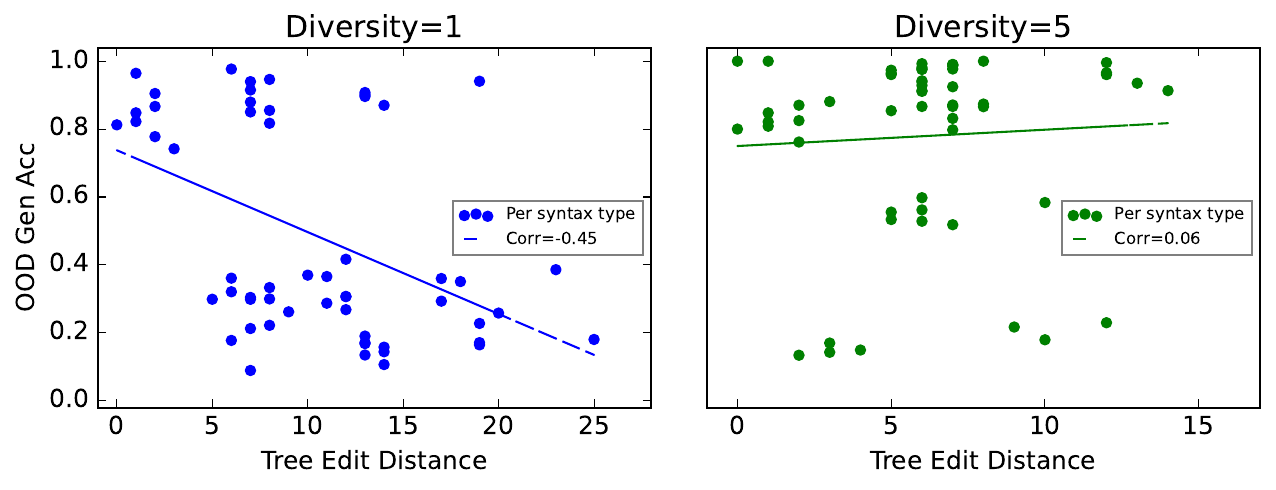}
    \caption{\textbf{OOD generalization vs. syntactic similarity to training data.} At low data diversity, model memorizes syntactic patterns and applies the hierarchical rule only on syntax structures similar to ones appeared in the training data. At high data diversity, model can extrapolates the hierarchical rule and can apply it even on unseen syntax structures that are dissimilar to training data.} 
    \label{fig:tree_edit_distance_performance}
\end{figure*}

We investigate model behavior when trained on data with limited diversity. By analyzing a model's generalization accuracy across different syntactic types, we aim to distinguish patterns indicative of either memorization or generalization.

\paragraph{Measuring data similarity} Building on the diversity measure from Section~\ref{sec:data_diveristy}, we now use Tree-Edit Distance (TED) as a measure of sentence similarity. As before, we first construct syntax trees using CFG rules, then calculate TED using the Zhang-Shasha Tree-Edit Distance algorithm \citep{Zhang1989-ma}. We define TED=0 for sentences that share the same syntax structure but differ only in vocabulary. This similarity measure allows us to quantify, for each sample in the OOD generalization set, the closest matching sentence type in the training data. In the memorization regime, where the model encounters only a few syntax types, we suspect it cannot extrapolate rules to syntactically distinct OOD sentences. In contrast, with a more diverse syntax exposure, rule extrapolation may enable the model to apply rules even to OOD sentence types.

\paragraph{Experiment} To verify our intuition about memorization and generalization, we train models on two variations of the QF data. In the first variation, the declaration-copying task has data diversity set to 1, meaning only one syntax type appears, and we specifically choose one with center embedding. In the second variation, the declaration-copying task has diversity set to 5, with all 5 types containing center embeddings. For both datasets, the question-formation task remains unchanged, consisting solely of right-branching sentences. 
For the diversity=1 dataset, we calculate TED for each unique syntax type in the OOD set against the single syntax type in the declaration-copying task. For the diversity=5 dataset, we compute TED between each OOD sample and the five syntax types in the declaration-copying task, taking the minimum. This TED score provides a measure of similarity between the OOD samples and those encountered during training.
Our goal is to determine, based on training with these datasets, which OOD syntax types the model applies the hierarchical rule to.

\paragraph{Result} In Figure~\ref{fig:tree_edit_distance_performance}, we visualize the final generalization accuracy for each OOD syntax type against its TED relative to the training data. When trained on low-diversity data (Figure~\ref{fig:tree_edit_distance_performance}, \textit{left}), generalization accuracy is negatively correlated with TED. For syntax types seen in the declaration-copying task (TED=0) and those similar to it, the model applies the hierarchical rule. However, for syntax types with high TED, the model’s behavior is random (25\%), indicating failure to follow any rule. As data diversity increases slightly (Figure~\ref{fig:tree_edit_distance_performance}, \textit{right}), generalization accuracy no longer correlates with TED, suggesting that once the model begins to extrapolate the hierarchical rule, it can apply this rule to a wider range of OOD syntax types.

\iffalse
\section{Additional Plots For Section~\ref{sec:stability}}
\todo{Update plot and write some explanation}
\begin{figure*}
    \centering
    \includegraphics[width=0.9\linewidth]{figures/intra_inter_inconsistency_3.png}
    \linebreak
    \includegraphics[width=0.5\linewidth]{figures/intra_inter_inconsistency_1.png}
    \linebreak
    \includegraphics[width=0.7\linewidth]{figures/intra_inter_inconsistency_2.png}
   
    \caption{\textbf{Three Different Inconsistency Behaviors.} \textit{Top:} Hierarchical Rule is Preferred. \textit{Middle:} Linear Rule is Preferred. \textit{Bottom:} No consistent rule is learned. }
    \label{fig:intra_inter_inconsistency_original}
\end{figure*}

\begin{figure}[h]
    \centering
    \includegraphics[width=1.0\textwidth]{figures/quest_decl_ratio.pdf}
    \caption{\textbf{Quest and Decl Ratio} Data size is standardized to 100k. We vary question and declaration ratios. For declarations, we also vary the the sentence typs mixes. \ns{for iclr we want to figure out why depth 1 doesn't teach a consistent rule at all. Also I don't understand how this plot would suggest depth one teaches no rule while it seems to give consistent linear rules based on figure 4} }
    \label{fig:quest_decl_ratio}
\end{figure}

\fi
\end{document}